\documentclass[preprint,12pt]{elsarticle}

\usepackage{amssymb}
\usepackage{amsmath}
\usepackage{arydshln}
\usepackage{multirow}
\usepackage[vlined, boxruled, linesnumbered]{algorithm2e}
\usepackage{makecell}
\usepackage{hyperref}
\hypersetup{
    colorlinks=true
}
\usepackage[nameinlink]{cleveref}
\usepackage{pdflscape}
\usepackage{longtable}
\usepackage{csquotes}
\usepackage[nolist,nohyperlinks]{acronym}
\usepackage{bookmark}
\usepackage{xr-hyper}
\begin{acronym}
    \acro{convqa}[ConvQA]{Conversational Question Answering} 
    \acro{llm}[LLM]{Large Language Model}
    \acro{io}[I/O]{Input Output}
    \acro{cot}[CoT]{Chain of Thought}
    \acro{mab}[MAB]{Multi-Agent Base}
    \acro{mad}[MAD]{Multi-Agent Debate}
    \acro{ar}[AR]{Agent Roundtable}
    \acro{rag}[RAG]{retrieval augmented generation}
    \acro{sc}[SC]{Self-Consistency}
    \acro{kpi}[KPI]{key performance indicator}
    \acro{io}[I/O]{Input Output}
    \acro{io}[I/O]{Input Output}
    \acro{io}[I/O]{Input Output}
    \acro{io}[I/O]{Input Output}
    \acro{io}[I/O]{Input Output}

\end{acronym}

\journal{Journal of Systems and Software}

\usepackage{multirow}%
\usepackage{mathrsfs}%
\usepackage{manyfoot}%
\usepackage{array}
\usepackage{amsfonts}
\usepackage{algorithmic}
\usepackage{graphicx}
\usepackage{textcomp}
\usepackage[dvipsnames]{xcolor}
\usepackage{listings}
\usepackage{sansmath}
\usepackage{hyperref}
\usepackage{tikz}
\usepackage{tcolorbox}
\usepackage{nicefrac}
\usepackage{balance}
\usepackage{lipsum}
\usepackage{subcaption}
\usepackage{float}
\usepackage{booktabs}
\usepackage{xspace}
\usepackage{tikz}
\usepackage{adjustbox}
\usepackage{subcaption}
\usepackage{amsmath}
\usepackage{wrapfig}
\usepackage{physics}
\usepackage{caption} 
\usepackage{lineno}
\usepackage[normalem]{ulem}
\usepackage{makecell}
\usepackage{xcolor}
\usepackage{colortbl}
\usepackage{graphicx}
\usepackage{tcolorbox}

\SetKwComment{Comment}{$\triangleright$\ }{}
\SetCommentSty{itshape}
\newboolean{showcomments}
\setboolean{showcomments}{true}
\ifthenelse{\boolean{showcomments}}
{
	\definecolor{myyellow}{RGB}{255, 228, 26}
	\definecolor{myblue}{RGB}{50, 50, 220}
	\newcommand{\nb}[2]{
		{\sf
			\fcolorbox{myyellow}{yellow}{\small\textbf{#1}}%
			{\color{myblue}\fontsize{12pt}{12pt}\selectfont\textbf{#2}}%
		}%
	}
}
{
	\newcommand{\nb}[2]{}
}

\newcommand\Abdullah[1]{\nb{Abdullah}{#1}}

\usepackage{soul}

\usepackage{xcolor} 
\usepackage{amsmath} 

\begin{document}

\begin{frontmatter}

\title{Benchmarking Contextual Understanding for In-Car Conversational Systems}
\author[hu,bmw]{Philipp Habicht}
\author[bmw,tum]{Lev Sorokin}
\author[bmw,tum]{Abdullah Saydemir}
\author[bmw,tum]{Ken Friedl}
\author[tum]{Andrea Stocco}

\affiliation[bmw]{organization={BMW Group},
            addressline={Petuelring 130}, 
            city={Munich},
            postcode={80809}, 
            country={Germany}}
\affiliation[hu]{organization={Humboldt University of Berlin},
            addressline={Unter den Linden 6}, 
            city={Berlin},
            postcode={10099}, 
            country={Germany}}
\affiliation[tum]{organization={Technical University of Munich},
            addressline={Arcisstraße 21}, 
            city={Munich},
            postcode={80333}, 
            country={Germany}}

\begin{abstract}\label{sec:abstract}
In-Car \ac{convqa} systems significantly enhance user experience by enabling seamless voice interactions. However, assessing their accuracy and reliability remains a challenge. 
This paper explores the use of \acp{llm} alongside advanced prompting techniques and agent-based methods to evaluate the extent to which \ac{convqa} system responses adhere to user utterances.
The focus lies on 
contextual understanding, the ability to provide accurate venue recommendations considering the user constraints and situational context.
To evaluate the utterance/response coherence using an \ac{llm}, we synthetically generate user utterances accompanied by correct but also modified failure-containing system responses.
We use input-output, chain of thought, self-consistency prompting, as well as multi-agent prompting techniques, with 13 reasoning and non-reasoning \acp{llm}, varying in model size and providers, from OpenAI, DeepSeek, Mistral AI, and Meta.

We evaluate our approach on a case study that involves a user asking for restaurant recommendations. 
The most substantial improvements are observed for small non-reasoning models when applying advanced prompting techniques, in particular, when applying multi-agent prompting.
However, non-reasoning models are significantly surpassed by reasoning models, where the best result is achieved with single-agent prompting incorporating self-consistency.
Notably, the DeepSeek-R1 model achieves the highest F1-score of 0.99 at a cost of 0.002 USD per request.
Overall, the best tradeoff between effectiveness and cost/time efficiency is achieved with the non-reasoning model DeepSeek-V3.

Our results demonstrate that \ac{llm}-based evaluations offer a scalable and accurate alternative to traditional human-based evaluations for benchmarking contextual understanding in \ac{convqa} systems. 
\end{abstract}


\end{frontmatter}


\section{Introduction}
\label{sec:intro}


\acf{convqa} systems are becoming increasingly important in various domains, especially in the automotive sector, where voice-controlled systems can significantly enhance safety and convenience.
These systems allow users to interact with vehicles using natural language, simplifying tasks such as navigation or controlling vehicle functions~\cite{mctear2024transforming, zaib2021conversationalquestionansweringsurvey}. 


One particular requirement of ConvQA systems is contextual understanding, the system's capability to align suggestions (e.g., restaurant recommendations) with real-time user context, such as location, time, or preferences.
For example, if a system provides a venue not matching the user's request (i.e., a restaurant that is too far away or closed), user trust may decline.
Given the increasing reliance on in-car systems, ensuring that \ac{convqa} systems work reliably is crucial for both user experience and safety~\citep{friedl2023incarethinkingincarconversational,guo2024mortarmetamorphicmultiturntesting,2020-Humbatova-ICSE,2020-Riccio-EMSE}. 

However, modern \ac{convqa} system are in general \ac{llm}-based~\cite{guo2024mortarmetamorphicmultiturntesting,2026-Dozono-ICSEW,2025-Guo-arxiv}, making them prone to hallucinate and produce incorrect responses to the user, an inherent limitation shared by most learning-based systems~\cite{RiccioEMSE20,2025-Baresi-ICSE,2025-Maryam-ICST}.
Consequently, \ac{convqa} systems need to be thoroughly tested before they are deployed in cars~\cite{2025-Giebisch-IV}. 
However, using human-based evaluation of conversational systems is time-consuming, expensive, and not scalable.
In addition,  automated metrics such as BLEU~\citep{papineni-etal-2002-bleu} or BERT~\citep{zhang2020bertscore}, although useful for basic evaluations, are not sufficiently accurate for complex conversational tasks involving contextually rich queries~\citep{guo2023evaluating}. 
Also, they require task-specific fine-tuning and are not generalizing well on unseen data, making them less practical to develop~\citep{peeters2023wdcproductsmultidimensionalentity}.
A promising alternative to overcome these challenges lies in using \acp{llm} for evaluation, as they demonstrate strong capabilities in understanding context and conversational dependencies without requiring fine-tuning. 

Moreover, existing research in which \acp{llm} evaluate general conversational quality has demonstrated strong alignment with human annotations~\citep{zheng2023judging, lin2023llmeval}.
Previous work by Friedl et al.~\cite{friedl2023incarethinkingincarconversational} has applied \acp{llm} to measure the accuracy of \ac{convqa} systems, providing valuable insights into their potential for automated evaluation.
Another study by Giebisch et al~\cite{2025-Giebisch-IV} focused on the factual relevance and consistency using a small set of selected models. 
Their work focuses on benchmarking of a judge which is used for evaluation of a \ac{rag} based conversational system.

Our study provides a comprehensive benchmarking framework, focusing in particular on contextual understanding. 
For the evaluation of judging methods in combination with \acp{llm}, we generate a manually validated dataset of textual inputs mimicking user inputs to \ac{convqa}, along with positive and negative responses including faults in context-related attributes.
We evaluate performance, cost, and time efficiency of 13 \acp{llm} of different sizes, as well as types regarding accessibility and reasoning capability.
In addition, we include 6 different prompting methods, including chain of thought prompting for step-by-step reasoning~\citep{wei2023chainofthoughtpromptingelicitsreasoning} and multi-agent prompting~\citep{du2023improvingfactualityreasoninglanguage, li2024improvingmultiagentdebatesparse}.
These methods have proven to better guide \acp{llm} in increasing their reasoning capabilities and significantly improve predictive accuracy~\cite {chen2024unleashingpotentialpromptengineering}. 



Our study makes the following contributions:
\begin{itemize}
    \item \textbf{Dataset}: We created a synthetic dataset of 600 recommendations for LLM-based judge evaluation. The dataset is human-validated and contains correct and incorrect recommendations to evaluate the abilities of the LLM-based judgment to understand misalignments in recommendations. The methodology behind the creation of dataset and a subset of the dataset are provided for replication: \href{https://github.com/judge-bench}{https://github.com/judge-bench}



    \item \textbf{Evaluation}: We perform an extensive evaluation of 13 \acp{llm} including reasoning as well as non-reasoning models with 6 prompting techniques and report the effectiveness and efficiency results.
    In addition, we evaluate which incorrect recommendation can be correctly identified by an LLM-based judgment and which cannot. To foster replicability, we provide prompts for every applied prompting technique in the appendix.

\end{itemize}

The main findings of our study are:
 
 \begin{itemize}
    \item Reasoning models outperform non-reasoning models in terms of F-1 scores.
    Regarding prompting techniques, a negligible effect is observed for reasoning models, while for non-reasoning models, the best improvement is identified for GPT-based models when using advanced prompting.
    Among the evaluated models, DeepSeek-R1 achieved the highest overall performance with an F-1 score of 0.99, while the best-performing non-reasoning model, DeepSeek-V3, reached an F-1 score of 0.98.
    Smaller reasoning models, such as o4-mini and o3-mini, consistently surpassed F-1 scores of 0.90, indicating that reasoning capabilities can compensate for smaller model sizes.
    \item Non-reasoning models show a high performance variability depending on the prompting strategy employed, with advanced techniques generally leading to improved outcomes.
    However, multi-agent prompting yields inconsistent results, particularly for reasoning models, performing similarly or even worse than single-agent-based prompting.
    \item In terms of efficiency, the Llama-405B model with self-consistency prompting represents the most resource-intensive configuration (approximately 50 seconds per evaluation), while Mistral with basic I/O prompting demonstrates the fastest response time (approximately 1 second).
    The most cost-efficient reasoning model is DeepSeek-R1, with an average cost of 0.002 USD per evaluation, while the most cost-efficient non-reasoning model is DeepSeek-V3 with 0.001 USD per evaluation offering at the same time the best trade-off between cost, latency, and performance.
\end{itemize}


The paper is structured as follows. 
\Cref{sec:related-work} presents related work. 
\Cref{sec:background} introduces our case study, followed by a description of applied prompting techniques in \Cref{sec:prompting}. In \Cref{sec:dataset} we present the dataset.
\Cref{sec:design} outlines the experimental design. 
\Cref{sec:results} presents the results and \Cref{sec:discussion} discusses their implications.
\Cref{sec:threats} addresses potential threats to validity.
\Cref{sec:conclusion} concludes the paper and outlines directions for future research. Additional details on the results and prompts are provided in the appendix.
\section{Related Work}\label{sec:related-work}


Generative systems are commonly evaluated using metrics based on word overlap or similarity, such as BLEU~\cite{papineni-etal-2002-bleu} or ROUGE~\cite{lin-2004-rouge}.
These metrics typically require gold-standard reference data, which may not always be available.
Additionally, they were originally designed for specific tasks: BLEU for machine translation and ROUGE for summarization.
These metrics impose rigid expectations on the generated text, offering limited tolerance for variation in phrasing or lexical choice.
Although they remain widely used, studies have shown that their correlation with human judgment is weak or negligible~\cite{zhang2020bertscore,rony-etal-2023-carexpert,novikova2017we}.

Several papers have outlined LLM-driven evaluation methods~\cite{chiang2023largelanguagemodelsalternative, lin2023llmevalunifiedmultidimensionalautomatic, honovich-etal-2021-q2} and their advantages~\cite{hosking2024humanfeedbackgoldstandard}, in particular for text summarization tasks. 
The idea of applying an LLM for evaluating other LLMs, called \textit{LLM-as-a-Judge}, has been presented initially by Zheng et al.~\cite{zheng2023judgingllmasajudgemtbenchchatbot}.
They have evaluated the judging capabilities of LLMs by using human curated question-answer pairs, including generic multi-turn conversations for categories such as math or text summarization with open-ended questions. 
Their evaluation shows that LLMs reach an agreement of more than 80\% with human experts.
However, their study is limited to a small selection of non-reasoning models, does not consider contextual understanding in the automotive context, and does not evaluate advanced prompting techniques.

In the automotive domain, Friedl et al.~\cite{friedl2023incarethinkingincarconversational} evaluated LLM-based judges for in-car-based conversational information retrieval. 
Multiple personas were created to let the LLM judge whether the system response fits to user question.
Both questions and answers were crafted by human experts. 
Their evaluation involving three LLM models with zero-shot prompting as well as multi-persona and max-vote prompting, shows that LLM-based judgment achieve up to 94\% agreement with human experts. 
However, the evaluation was only performed on follow-up question answering, implicit understanding, and handling harmful user inputs. 

Giebisch et al.~\cite{2025-Giebisch-IV} evaluated LLM-based-judgment regarding the factual correctness in the in-car context.
They have shown that LLM-based judgment achieves up to 90\% agreement with human experts. 
However, their study is limited to evaluating requests for the properties of factual correctness and consistency for a \ac{rag}-based system~\cite{rony-etal-2023-carexpert}. 
Our work, on the other hand, focuses on the contextual understanding and provides an extensive study including small but also large scale conventional as well as reasoning models. 
In addition, it is independent of the underlying conversational system technique.

Conversational datasets such as CoQA~\cite{reddy2019coqa}, MMDialog~\cite{feng2022mmdialog}, and VACW~\cite{siegert-2020-alexa}, and datasets such as MultiWOZ2.2~\cite{zang2020multiwoz}, KVERT~\cite{eric2017keyvalue}, which in particular focus on navigational requests and recommendations, do exist. However, these datasets do not incorporate all contextual parameters such as time, cost, and location in a single request, nor simulate incorrect system responses which are required for our contextual understanding benchmark.



\section{Motivation}\label{sec:background}

In the following, we motivate our work by introducing our case study of benchmarking conversational question answering systems (\ac{convqa}).

\ac{convqa} systems are a branch of conversational AI designed to understand and respond to user input in multi-turn dialogues.
These systems incorporate functionalities such as information retrieval, API calls, and web searches to handle complex user interactions~\cite{mctear2024transforming, zaib2021conversationalquestionansweringsurvey}.
\ac{convqa} systems are applicable across various domains, with one significant area being the automotive industry. 
In vehicles, these systems facilitate natural language interactions with core functions such as navigation, access to the car manual, and control of car features like audio settings or climate control~\cite{rony2023carexpertleveraginglargelanguage, 2025-Giebisch-IV}.

\begin{figure}[t]
    \centering
    \includegraphics[trim=27 140 7 100, clip, scale=0.42]{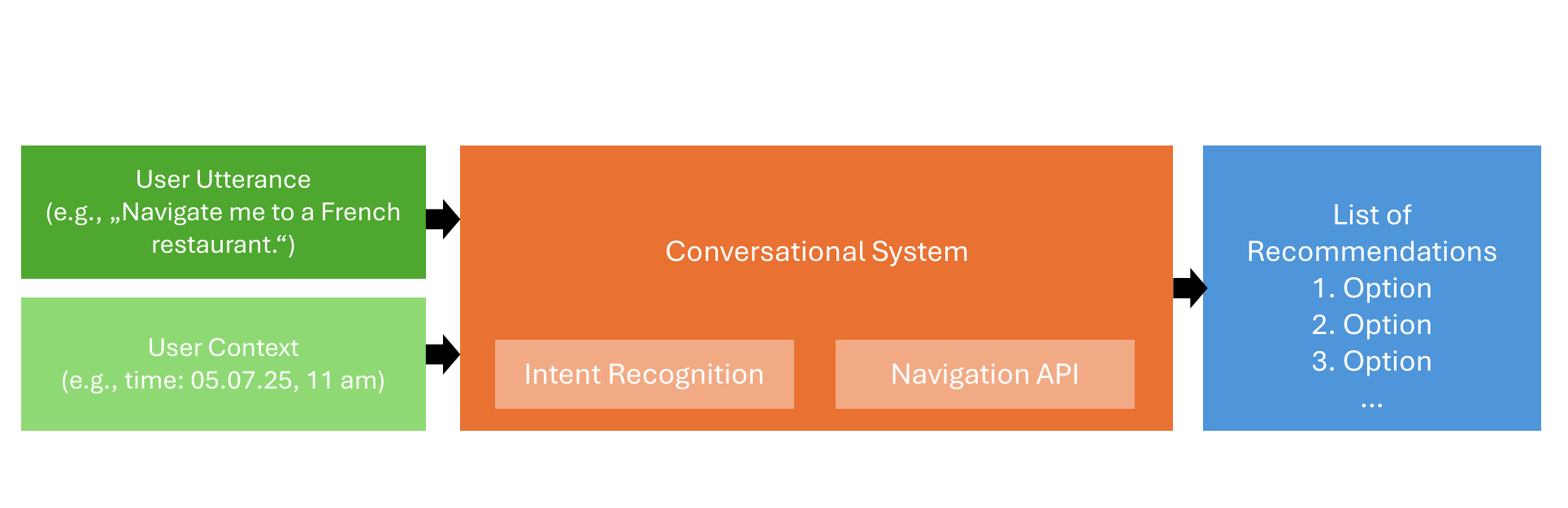}
    \caption{Example of a \ac{convqa} for navigation purposes. The left hand side depicts the system inputs, while the right hand side depicts its outputs with venue recommendations.}
    \label{fig:convqa}
\end{figure}

\Cref{fig:convqa} illustrates a \ac{convqa} system integrated into a vehicle for navigation purposes.

An example workflow in processing the user request is as follows: first, the system captures the user input, which is, in general, provided in the form of a speech utterance. In the next step, the speech utterance is converted into text, and in following analyzed by the intent recognizer. The intent recognizer tries to identify an appropriate tool to process the request further. In our example, the request is handled over to the navigational tool, which collects first contextual data such as the users location and the time of the request, followed by calling the navigational API, which provides location information of relevant venues. The system then selects some nearby venues, considering the user preferences provided in the request.

For the remainder of this paper, we denote the user utterance combined with its contextual data such as location, data and time as \textit{user block}. 
We denote the response that the system generates as \textit{system block}. The response contains parametrized and detailed output of venue information, including attributes such as \textit{name}, \textit{location}, \textit{cost} or \textit{opening hours}, \textit{rating}, and \textit{cuisine}.

\begin{figure}[t]
    \centering
    \includegraphics[trim=0 26 0 0, clip,width=0.95\textwidth]{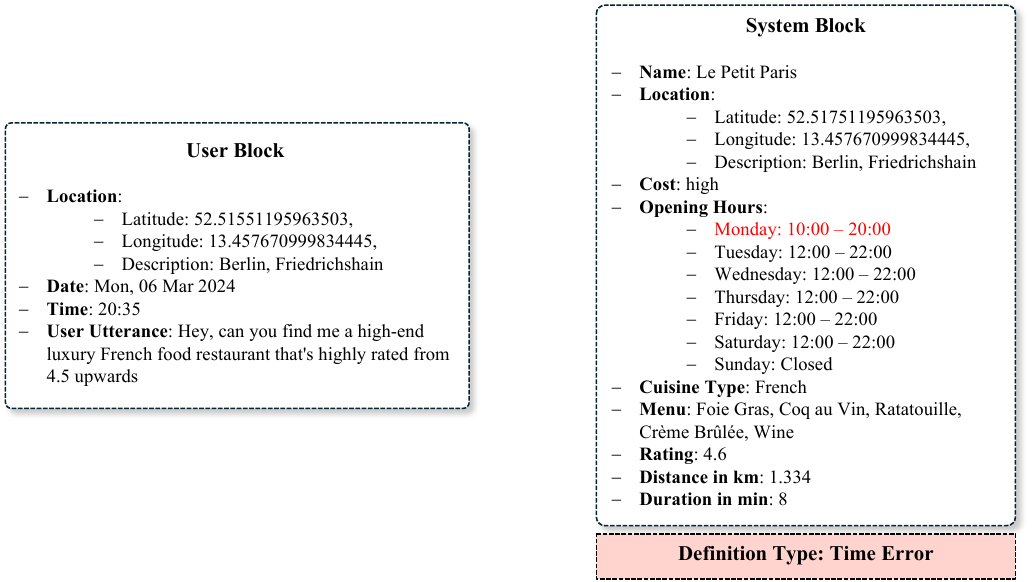}
    \caption{Information passed to the judge under test: the user block with the request for a high-end French restaurant at 20:35, and an example of a system response, where the system recommends a non-suitable venue, where opening hours end on Monday at 20:00.}
    \label{fig:con_example}
\end{figure}

Examples of a user block and system block are shown in \Cref{fig:con_example}. In this example, all venue information in the system block align with the users request, except for the opening times. I.e., the proposed venue is closed at the time when the user sends the request.

The goal of this study is to evaluate LLM-based judgment techniques that assess the quality of a system response.
Specifically, how well the system block aligns with a given user block.

\section{Prompting Techniques}
\label{sec:prompting}
In our study, we evaluate different prompting techniques when applying LLMs for judgment. Prompt engineering has become a crucial method for enhancing the performance of \ac{llm}.
By providing task-specific instructions, prompting allows models to adapt to various downstream tasks without modifying their core parameters. 
Instead of retraining or fine-tuning the model, prompts are designed to give context and steer LLMs toward desired behaviors~\citep{sahoo2024systematicsurveypromptengineering, chen2023programthoughtspromptingdisentangling}.
As shown in previous studies~\cite{friedl2023incarethinkingincarconversational, 2025-Giebisch-IV}, the prompting method can have a significant impact on the performance of a model. An overview of all prompting techniques used in this study is provided in \Cref{tab:prompting-techniques} and explained in the following in detail.

\begin{table}[t]
\centering
\footnotesize
\setlength{\tabcolsep}{6pt} 
\renewcommand{\arraystretch}{1.2} 
\caption{Overview of applied prompting techniques, with brief descriptions and variation used (e.g., number of agents or outputs).}
\begin{tabular}{lp{4cm}l}
\toprule
\textbf{Name} & \textbf{Description} & \textbf{Variation} \\
\midrule
\textbf{Input-Output (I/O)} & Passing one prompt to one model without examples & 1 \\
\hline
\textbf{Chain of Thought (CoT)} & Give n examples for reasoning about the task alongside the prompt & 1, 3, 5 \\
\hline
\textbf{Self-Consistency (SC)} & Multiple CoT reasoning paths; selection of most frequent verdict & 3, 5 \\
\hline
\textbf{Multi-Agent Base (MAB)} & Agents (same LLM) with distinct roles discuss verdict for one round & 1 \\
\hline
\textbf{Multi-Agent Debate (MAD)} & Agents (same LLM) with distinct roles discuss verdict until agreement, for a given max. number of rounds & 3\\
\hline
\textbf{Agent Roundtable (AR)} & Agents of different LLMs deliberate until agreement, for a given max. number of rounds & 3 \\
\hline
\end{tabular}
\label{tab:prompting-techniques}
\end{table}



\subsection{Input-Output Prompting}

\ac{io} prompting, as outlined in~\cite{liu2021pretrainpromptpredictsystematic}, is the most standard technique for prompting \acp{llm}.
Given an input or prompt \( x \), the goal is to generate a corresponding output \( y \).
This method leverages the autoregressive properties of transformer models, which sequentially map \( x \) to \( y \) by estimating the conditional probability distribution:
\[
\mathbb{P}(y \mid x)
\]
To generate the actual output \( y^* \), the model selects the most probable output by maximizing:
\[
y^* = \arg\max_{y} \mathbb{P}(y \mid x)
\]
For a practical example of \ac{io} prompting and the prompt templates used, please refer to \Cref{fig:con_input_output_prompt} in the Appendix. 

\subsection{Chain of Thought}

\ac{cot} prompting is a technique developed to enhance the reasoning capabilities of language models by guiding them to generate structured, step-by-step thought processes~\citep{wei2023chainofthoughtpromptingelicitsreasoning}.
Unlike \ac{io} prompting, which directly maps an input \( x \) to an output \( y \), \ac{cot} prompting encourages the LLM to produce intermediate reasoning steps \(z = (z_1, z_2, \ldots, z_M) \) that lead to the final answer \( y \).
Each step \( z_i \), where \( i = 1, \dots, M \), represents a logical inference or thought that connects \( x \) to \( y \), effectively mimicking human problem-solving methods by breaking down complex tasks into manageable sub-steps. 

To implement \ac{cot} prompting, we employ few-shot (n-shot) prompting by providing the model with \( N \) examples, also known as shots.
Each example  \( e_j \), for \( j = 1, \dots, N \), consists of an input \( x_j \), a corresponding sequence of reasoning steps $z_j = (z_{j_1}, z_{j_2}, \ldots, z_{j_{M}}) $, and an output \( y_j \).
These examples guide the model in generating reasoning steps for new inputs.

Given an input \( x \), the model aims to generate appropriate reasoning steps \( z \) and produce the final output \( y \), leveraging the provided examples \(E = \{e_1, e_2, \ldots, e_N\} \). 
The probabilistic formulation of this process is expressed as:
\[
\mathbb{P}(y, z \mid x, E) = \mathbb{P}(z \mid x, E) * \mathbb{P}(y \mid x, z, E)
\]
To generate the final output, the LLM selects the output that maximizes the joint probability:
\[
(y^*, z^*) = \arg\max_{y, z} \mathbb{P}(y, z \mid x, E)
\]
\acs{cot} prompting has been shown to significantly improve the accuracy of language models in tasks that require common sense, mathematical, and symbolic reasoning compared to the more straightforward \ac{io} prompting.
By incorporating a few-shot examples, the model learns to generalize the reasoning process to new inputs, enhancing its problem-solving capabilities.
For a practical example of \ac{cot} prompting and the prompt templates used, please refer to \Cref{fig:con_cot_prompt} in the Appendix. 

\subsection{Self-Consistency}

\ac{sc} builds upon \ac{cot} prompting by addressing the variability and randomness inherent in the reasoning process of LLMs.
Due to the non-deterministic nature and the diversity of possible reasoning paths, the model may produce different outputs \( y \), when prompted identically.
To mitigate this, \ac{sc} involves independently sampling multiple reasoning paths \(\{z_1, z_2, \ldots, z_n\} \) for a given input \( x \), effectively generating a diverse set of candidate answers \(\{y_1, y_2, \ldots, y_n\} \).
The underlying assumption is that correct reasoning processes are more likely to converge on the same answer, whereas incorrect or flawed reasoning will result in a wider variety of answers.
In practice, for each sampled reasoning path \( z_i \), the model produces a corresponding output \( y_i \).
The process can be formulated as:
\[
(y_i, z_i) = \arg\max_{y} \mathbb{P}(y, z \mid x, E)
\]
where \( E \) represents the set of few-shot examples used in \ac{cot} prompting.
Here, \( y_i \) and \( z_i \) are jointly sampled from the probability distribution conditioned on the input \( x \) and the examples \( E \).
After obtaining multiple outputs, the final answer is determined by aggregating these outputs and selecting the most frequent among them: 
\[
y^* = \text{mode}\left(y_1, y_2, \ldots, y_n\right)
\]
This method improves the accuracy of language models in tasks involving arithmetic, commonsense reasoning, and other complex problem-solving scenarios, as it mitigates the risks associated with a single, potentially flawed reasoning path~\citep{wang2023selfconsistencyimproveschainthought}.

\subsection{Multi-Agent Systems}
Building on these fundamentals, agentic methods were developed where agents with specific characteristics are able to collaborate to make decisions.
Inspired by the \textit{society of minds} concept~\citep{minsky1986society}, this approach highlights agent communication for tackling complex tasks~\citep{wang2024unleashingemergentcognitivesynergy}. 
In the following, we will explain how we have applied these methods in our research, whereby there may be deviations from the original implementations.

\subsubsection{Multi-Agent Base}

To further enhance decision making capabilities, we adopted a multi-agent approach as outlined in~\cite{friedl2023incarethinkingincarconversational}.
In this method, here called \ac{mab}, a set of agents tailored to the \acp{kpi} was defined.
These agents independently make decisions based on the input, and the final outcome is determined by aggregating their responses.

Let \( x \) be the given input, and let \( \{A_1, A_2, \ldots, A_M\} \) denote the set of \( M \) agents.
Each agent \( A_i \) is characterized by its own persona in a prompt \( P_i \), which influences its interpretation and processing of the input \( x \).
Furthermore, each agent \( A_i \) generates reasoning steps \( z_i \) and a corresponding output \( y_i \), resulting in: 
\[
\mathbb{P}_{A_i}(y, z \mid x) = \mathbb{P}(y, z \mid x, P_i)
\]
Each agent \( A_i \) generates its reasoning and outputs by maximizing:
\[
(y_i, z_i) = \arg\max_{y, z} \mathbb{P}_{A_i}(y, z \mid x)
\]
The set of responses from all agents is \(\{(y_1, z_1), (y_2, z_2) \ldots, (y_M, z_M)\} \).
To arrive at the final decision \( y^* \), we aggregate the outputs \( y_i \) using mode:
\[
y^* = \text{mode}\left(y_1, y_2, \ldots, y_M\right)
\]
The rationale is that leveraging agents with diverse perspectives leads to sampling from different parts of the probability distribution, resulting in varying responses.
Further, when agents independently reach the same conclusion, it boosts confidence in the decision.
Aggregating their responses creates a more balanced and robust outcome by reducing individual biases or errors.
Prompt templates and examples for the agents will be shown and further discussed in the next chapter.

\subsubsection{Multi-Agent Debate}
The agentic approach by~\cite{li2024improvingmultiagentdebatesparse}, called \ac{mad} in this study, extends the \ac{mab} method by allowing multiple rounds of discussion among agents.
Instead of each agent independently producing a decision, agents are able to collaborate over several rounds, refining their responses based on the insights of others, if no agreement was found earlier.
This iterative process leads to improved accuracy, reduced bias, and better handling of uncertainties, as agents collectively address ambiguities and refine their decisions~\citep{du2023improvingfactualityreasoninglanguage}.
\ac{mad} consists of three phases: 

\paragraph{Phase 1: Initial Response Generation}
Let \( x \) be the input, and let  $\{$$A_1$, $A_2$, $\ldots$, $A_M$$\}$ denote the set of \( M \) agents, each defined by a specific persona in a prompt \( P_i \). 
Based on its unique characteristics, each agent \( A_i \) generates initial reasoning steps \( z_i^{(0)} \) and an initial response \( y_i^{(0)} \):
\[
(y_i^{(0)}, z_i^{(0)}) = \arg\max_{y} \mathbb{P}_{A_i}(y \mid x)
\]

\paragraph{Phase 2: Multi-Round Discussion}
If no agreement is reached after the initial round, agents exchange their reasoning and outputs and engage in discussions.
In each subsequent round \( r = 1,2, \ldots, R \), each agent \( A_i \) refines its response by considering the input \( x \), its own previous reasoning and response \( (y_i^{(r-1)}, z_i^{(r-1}) \), and the reasoning and responses from all other agents in the previous round \(\{(y_j^{(r-1)}, z_j^{(r-1)}) \mid j = 1, 2, \dots, M; j \neq i\}\), resulting in:
\[
(y_i^{(r)}, z_i^{(r)}) = \arg\max_{y, z} \mathbb{P}_{A_i}(y, z \mid x, y_i^{(r-1)}, z_i^{(r-1)}, y_{j}^{(r-1)}, z_{j}^{(r-1)})
\]
This process continues for up to \( R \) rounds or until a consensus is reached.
Consensus is achieved when all agents \( A_i \) agree on the same output:
\[
y_1^{(r)} = y_2^{(r)} = \dots = y_M^{(r)} = y^*
\]

\paragraph{Phase 3: Final Decision}
If consensus is not reached after \( R \) rounds, the final decision \( y^* \) is determined by majority voting among the responses in the last round:
\[
y^* = \text{mode}(y_1^{(R)}, y_2^{(R)}, \dots, y_M^{(R)})
\]
\ac{mad} approach operates on the hypothesis that through iterative discussions, agents can converge toward a more accurate decision by learning from reasoning of others.
Engaging in multiple rounds allows agents to address ambiguities, correct misunderstandings, and collectively refine their responses.

\subsubsection{Agent Roundtable}

\ac{ar} approach by~\cite{chen2024reconcileroundtableconferenceimproves} differs from \ac{mad} method by two key components.
First, instead of employing multiple agents within a single model, it utilizes \ac{cot} prompting across different LLMs $L_i$, such as GPT-4o, Mistral-Nemo, and DeepSeek-R1.
This mitigates the risks of inherent model biases, limited knowledge scopes, and the absence of external feedback that could arise if all answers were generated by the same data and model architectures.
Second, it utilizes uncertainty confidence estimation prompts, as introduced by ~\cite{tian2023justaskcalibrationstrategies}, to allow each agent to assess and express its confidence in the correctness of its response, enabling more confident responses to guide the discussion and making it easier to resolve uncertainties and reach consensus.
\ac{ar} method also consists of three phases:

\paragraph{Phase 1: Initial Response Generation}
Given the input \( x \) and a set of few-shot examples \( E \) for \ac{cot} prompting, each \ac{llm} \( L_i \) generates initial reasoning steps \( z_i^{(0)} \), an initial response \( y_i^{(0)} \) and an associated confidence level \( p_i^{(0)} \in [0, 1] \) of how confident the \ac{llm} is about its correctness of its decision \( y \):
\[
(y_i^{(0)}, z_i^{(0)}, p_i^{(0)}) = \arg\max_{y, z, p} \mathbb{P}_{L_i}(y, z, p \mid x, E)
\]

\paragraph{Phase 2: Multi-round Discussion}
If no agreement is reached after the initial round, the \acp{llm} exchange their reasoning, outputs, and confidence levels.
In each subsequent round \( r = 1,2, \ldots, R \), each LLM \( L_i \) refines its reasoning, output and confidence score:
$$(y_i^{(r)}, z_i^{(r)}, p_i^{(r)}) = \arg\max_{y, z, p} \mathbb{P}_{L_i}(y, z, p \mid x, E, y_i^{(r-1)}, z_i^{(r-1)}, p_i^{(r-1)}, y_{j}^{(r-1)}, z_{j}^{(r-1)}, p_{j}^{(r-1)})$$

This process continues for up to \( R \) rounds or until a consensus is reached, defined as the point when all \acp{llm} \( L_i \) agree on the same output:
\[
y_1^{(r)} = y_2^{(r)} = \dots = y_M^{(r)} = y^*
\]

\paragraph{Phase 3: Final Decision}
At the end of the final round \( R \), the final decision \( y^* \) is determined using a confidence-weighted voting scheme.
First, each \acp{llm} confidence \( p_i^{(R)} \) is calibrated using a function \( f(p_i^{(r)}) \). 
Calibrated confidence scores are used as weights to compute the final answer:
\[
y^* = \arg\max_y \sum_{i=1}^{M} f(p_i^{(r)}) \cdot \mathbf{1}(y_i^{(R)} = y)
\]
where \( y \) is a distinct output generated by any of the \acp{llm} (e.g. true or false), \( p_i^{(r)} \) is the original confidence of the \acp{llm} \( L_i \), and \( f(p_i^{(r)}) \) is the calibrated confidence defined as:
\[
f(p_i^{(r)}) = \begin{cases} 
1.0 & \text{if } p_i^{(r)} = 1.0 \\
0.8 & \text{if } 0.9 \leq p_i^{(r)} < 1.0 \\
0.5 & \text{if } 0.8 \leq p_i^{(r)} < 0.9 \\
0.3 & \text{if } 0.6 \leq p_i^{(r)} < 0.8 \\
0.1 & \text{otherwise} 
\end{cases}
\]
Transformation of \ac{llm} confidence levels is necessary because they often struggle with accurately interpreting numerical relationships in rankings; therefore, calibration helps aligning their confidence scores with their actual performance. 
\section{Dataset Generation}
\label{sec:dataset}

To evaluate \acp{llm} for contextual understanding, we have developed a dataset in collaboration with our industrial partner BMW. The dataset consists of 100 synthetically generated user requests (i.e., user blocks) and 600 system blocks simulating a conversational system's response.  
Each user block combines natural language with contextual metadata such as time, date, and location, as well as semantic preferences regarding cuisine, price, and rating, simulating a user's input. 

For every user block, the dataset includes one correct system block (recommendation) where all parameters are aligned and five incorrect recommendations with misalignments based on induced errors. By having misaligned recommendations, the goal is to validate whether the \acp{llm} under evaluation can reliably detect discrepancies in the responses, which is relevant when applying the judging technique later for the detection of failures in conversational systems. In the following, we explain in detail how the user and system blocks are generated automatically.

\begin{algorithm}[t]
\DontPrintSemicolon
\footnotesize
\SetKwInOut{Input}{Input}
\SetKwInOut{Output}{Output}

\Input{
    $\mathcal{L}$: List of 10 urban locations (in Berlin, Munich) \newline
    $\mathcal{D}$: Set of calendar dates from 2024 \newline
    $\mathcal{T}$: Time range between 08:00 and 22:00 \newline
    $\mathcal{C}$: Set of 20 cuisine types with 5 lexical variants each \newline
    $\mathcal{K}_{\text{cost}}$: Cost categories (low, medium, high) with 15 paraphrases each \newline
    $\mathcal{R}_{\text{phrases}}$: Expressions for ratings above 3.5 (e.g., "above 3.8", "at least 4.5") 
}
\Output{
    $\mathcal{U}_{\text{blocks}}$: Set of 100 user blocks 
}

\BlankLine
$\mathcal{U}_{\text{blocks}} \gets \emptyset$\;

\For{$i \gets 1$ \KwTo $100$}{
    $l \gets \texttt{sampleUniform}(\mathcal{L})$\;
    $d \gets \texttt{sampleUniform}(\mathcal{D})$\;
    $t \gets \texttt{sampleUniform}(\mathcal{T})$\;

    $(c, c_{\text{lex}}) \gets \texttt{sampleCuisineWithKeyword}(\mathcal{C})$\;
    $k_{\text{cost}} \gets \texttt{sampleParaphrase}(\mathcal{K}_{\text{cost}})$\;

    $r \gets \texttt{selectRatingPhrase}(c, k_{\text{cost}}, \mathcal{R}_{\text{phrases}})$\;

    $u_{\text{utt}} \gets \texttt{generateUtterance}(c_{\text{lex}}, k_{\text{cost}}, r)$\;

    $\texttt{ctx} \gets \texttt{formatContext}(l, d, t, c, k_{\text{cost}}, r)$\;
    $u_{\text{block}} \gets \texttt{merge}(u_{\text{utt}}, \texttt{ctx})$\;

    $\mathcal{U}_{\text{blocks}} \gets \mathcal{U}_{\text{blocks}} \cup \{u_{\text{block}}\}$\;
}

\KwRet{$\mathcal{U}_{\text{blocks}}$}
\caption{Generation of user blocks with contextual information}
\label{algo:user-block-generation}
\end{algorithm}

\subsection{User Block Generation}\label{sec:user-block-generation}

To generate diverse in-car user-system interactions for contextual understanding, we follow a structured procedure as outlined in~\Cref{algo:user-block-generation}. The generation begins by initializing an empty list of user blocks (line 1) and iterating 100 times to create a new user block. To generate a new user block, the algorithm samples uniformly a location from a predefined list of ten urban areas across Berlin and Munich (line 3).

Each location is represented by a coordinate pair and a district label (e.g., “Prenzlauer Berg, Berlin”) to allow for downstream geographic relevance evaluation. Next, a calendar date \textit{d} is randomly drawn from a list of dates $\mathcal{D}$ including dates in the year 2024 (line 4). The algorithm then samples a time \text{t} uniformly from the range 08:00 to 22:00 (line 5), covering dining hours during which users request restaurant recommendations while driving.

After establishing the contextual frame, the algorithm proceeds to define the user's preferences.
It samples a cuisine type \textit{c} from a set of 20 international options (e.g., Italian, Korean, Brazilian), and one of five lexical variants $c_{\text{lex}}$ associated with the selected cuisine (line 6).
For instance, \textit{Sushi} or \textit{Ramen} might be selected as a keyword for the Japanese category, to ensure lexical diversity.
Further, a cost level (low, medium, or high) is defined next (line 7), along with a corresponding paraphrased expression that reflects natural user language, such as \textit{rock-bottom prices} or \textit{luxurious prices}.

We pass both the cost level and the paraphrased expression to GPT-4 to generate a rating phrase from a predefined set of expressions (e.g., \textit{above 3.8}) (line 8). This conditional selection targets to maintain logical consistency, avoiding unrealistic combinations such as \textit{dirt-cheap} with \textit{five-star rating}. All ratings are restricted to values above 3.5, aligned with BMW's internal quality classification standards. With the user preferences defined, we pass the information to GPT-4 to generate an utterance $u_{\text{utt}}$ that combines the selected cuisine keyword, cost phrase, and rating expression (line 9).

In the remaining steps of the algorithm, we combine both the user utterance with contextual information into a user block and store it in a list (line 11-12). An example of such a generated utterance is: 
\textit{“Hey, can you find me a high-end luxury French restaurant with a rating of at least 4.5?”}. After completing all 100 iterations, the algorithm returns the set of user blocks.

\begin{algorithm}[t]
\DontPrintSemicolon
\footnotesize
\SetKwInOut{Input}{Input}
\SetKwInOut{Output}{Output}

\Input{
    $\mathcal{U}$: Set of user blocks
}
\Output{
    $\mathcal{D}_{\text{pos}}$: Set of correct recommendations
}

\BlankLine
$\mathcal{D}_{\text{pos}} \gets \emptyset$\;
\ForEach{$u \in \mathcal{U}$}{
    $p \gets \texttt{constructPrompt}(u, \texttt{fully\_aligned})$\;
    $r \gets \texttt{GPT4.generate}(p)$\;
    $\mathcal{D}_{\text{pos}} \gets \mathcal{D}_{\text{pos}} \cup \{(u, r)\}$\;
}
\KwRet{$\mathcal{D}_{\text{pos}}$}
\caption{Generation of correct recommendations}
\label{algo:positive-case-generation}
\end{algorithm}

\subsection{System Block Generation} \label{sec:system-block-generation}
For each user block, we generate six system blocks representing restaurant recommendations. One of these recommendations is a \textit{positive case} (i.e., a fully aligned recommendation), while the remaining five recommendations are \textit{error cases}, each containing exactly one misalignment in one of the following dimensions: location, time, cuisine, cost, or rating. All other parameters are kept identical to isolate the effect of the specific discrepancy. The generation process is illustrated in~\Cref{algo:positive-case-generation} for positive cases and in~\Cref{algo:error-case-generation} for error cases.

\paragraph{Correct Recommendations} 
In~\Cref{algo:positive-case-generation}, the process begins by initializing an empty list of aligned system blocks $\mathcal{D}_{\text{pos}}$ (line~1).
For each user block $u \in \mathcal{U}$ (line~2) (i.e., the ones generated in \autoref{algo:user-block-generation}), a prompt $p$ is constructed using a template where all context parameters do fully align (line~3). GPT-4 is then prompted to generate a recommendation (line~4), which is stored together with its corresponding user block in (line~5). After processing all user blocks, the algorithm returns the complete set of positive blocks (line~6).

\paragraph{Incorrect Recommendations} To generate an incorrect recommendation (\Cref{algo:error-case-generation}), the system iterates over the same user block generated with \autoref{algo:user-block-generation}, serving as the reference for controlled modification. In particular, the algorithm iterates over each error type and generates for each error type an error-specific prompt to introduce exactly one targeted error, e.g., by asking for a different cuisine than given in the request (line~4). The prompt is than passed to GPT-4 to generate a faulty recommendation $r_e$ (line~6). 

Location errors are treated in a specific way: we perform a post-processing
and make an external call to the Mapbox API~\cite{mapboxapi} to compute the driving time between the original user location and the generated restaurant location (line~8).
If the travel time is below the 15-minute threshold, a new recommendation is generated until the constraint is satisfied. All error-specific recommendations are then stored and returned at the end of the algorithm (line~12). For instance, \Cref{fig:con_example} illustrates a user and system block pair with a time error generated with this approach. 
For each error case, all parameters except for the error-related parameter remain identical to the corresponding positive case, to ensure that we can evaluate later the judgments performance sensitivity to specific error categories.

Finally, after automatic generation, we passed all system blocks to three domain experts for review to validate the correctness of the prompts. From in total of 600 generated requests, approximately 30\% samples required manual refinement due to inaccuracies in GPT-4 generations because of incorrectly aligned or incorrectly misaligned parameters.

\begin{algorithm}[t]
\DontPrintSemicolon
\footnotesize
\SetKwInOut{Input}{Input}
\SetKwInOut{Output}{Output}

\Input{
    $\mathcal{U}$: Set of user blocks \newline
    $\mathcal{D}_{\text{pos}}$: Fully aligned recommendations \newline
    $E = \{\texttt{location}, \texttt{time}, \texttt{cuisine}, \texttt{cost}, \texttt{rating}\}$
}
\Output{
    $\mathcal{D}_{\text{err}}$: Dictionary of error-specific recommendation sets
}

\BlankLine
\ForEach{$u \in \mathcal{U}$}{
    $r_{\text{base}} \gets \texttt{lookup}(\mathcal{D}_{\text{pos}}, u)$\;
    \ForEach{$e \in E$}{
        $p_e \gets \texttt{constructPrompt}(u, e)$\;
        \Repeat{
            $r_e \gets \texttt{GPT4.generate}(p_e)$\;
            \If{$e = \texttt{location}$}{
                $t \gets \texttt{MapboxAPI.estimateTravelTime}(u.\texttt{location}, r_e.\texttt{location})$\;
            }
        }{
            $(e \ne \texttt{location})$ \textbf{or} $t > 15$
        }
        $\mathcal{D}_{\text{err}}[e] \gets \mathcal{D}_{\text{err}}[e] \cup \{(u, r_e)\}$\;
    }
}
\KwRet{$\mathcal{D}_{\text{err}}$}
\caption{Generation of recommendations with errors}
\label{algo:error-case-generation}
\end{algorithm}

\section{Experimental Design}
\label{sec:design}
In this section, we describe the experimental setup for the evaluation of LLM-based judgement for benchmarking contextual understanding. We describe
the research questions, the language models and prompting techniques under test, the evaluation metrics, and the deployment.

\subsection{Research Questions}



\noindent\textbf{RQ\textsubscript{0} (Data Validation):} \textit{What is the validity rate of the benchmarking dataset?}

We evaluate the validity of our benchmarking dataset by performing a user study and passing the system and user blocks to humans who have not been involved in this paper.
By evaluating both the validity of in and outputs, and the relation between system and user blocks, we are able to evaluate whether the dataset can be used four LLM-based judge evaluation.

\noindent\textbf{RQ\textsubscript{1} (Effectiveness):} \textit{What combination of LLM and method is the most effective for contextual understanding in navigational requests?}

By answering this question, we want to evaluate how well LLMs can identify contextually incorrect as well correct system responses for a given user request. 
    
\noindent\textbf{RQ\textsubscript{2} (Efficiency):} \textit{What are the trade-offs between processing time and costs when using different LLMs and methods to benchmark contextual understanding?}

In the second research question, we want evaluate the runtime and costs which are required to perform an evaluation of contextual understanding using different model and method configurations. Runtime and costs are specifically relevant when a large number of tests needs to be executed when testing a ConvQA.

\noindent\textbf{RQ\textsubscript{3} (Failure):} \textit{On which misaligned recommendations do LLMs exhibit failures in contextual understanding?}

By answering the third research question, we want to understand which errors in recommendations lead to incorrect judgment results when applying LLMs for evaluation. 
 


\subsection{Language Models}

For the evaluation of different \ac{llm}-based judges using the methods introduced in the previous chapter, we selected diverse \acp{llm} varying in size, accessibility, type, date, and costs.
\Cref{tab:llms_overview} provides a detailed overview of the models used in this study, highlighting their key specifications and associated costs.
Note that OpenAI charges a discounted rate for batched inputs and DeepSeek charges less for queries during the night.
We report the baseline numbers and did not resort to either of these.

\begin{table}
    \centering
        \caption{LLMs used in this study. Note that pricing rates are from the time of the runs and might be different from current rates. Reasoning models are denoted with (R). Token pricing is as of July 2025 and may differ from current rates.}
    \label{tab:llms_overview}
    \begin{adjustbox}{width=\textwidth}
        \begin{tabular}{lcccccc}
        \toprule
            \textbf{Model Name} & \textbf{Company} & \textbf{\thead{Context\\ Window}} & \textbf{\thead{Knowledge \\ Cut-Off}} & \textbf{\thead{Number of\\ Parameters}} & \textbf{\thead{Cost / 1M \\ Input Tokens}}  & \textbf{\thead{Cost / 1M \\ Output Tokens}}  \\
        \midrule

            GPT-3.5 Turbo & \multirow{7}{*}{OpenAI} & 16K & Sep 2021 & \multirow{7}{*}{unknown} & \$0.50 & \$1.50 \\
            GPT-4 Turbo   &  & 128K & Dec 2023 &  & \$10.00 & \$30.00  \\
            GPT-4o        &  & 128K & Oct 2023 &  & \$5.00 & \$15.00 \\
            GPT-4.1       &  & 1M   & Jun 2024 &  & \$2.00 & \$8.00  \\
            o3-mini (R)      &  & 200K & Oct 2023 &  & \$1.10 & \$4.40 \\
            o3 (R)            &  & 200K & Jun 2024 &  & \$10.00 & \$40.00 \\
            o4-mini (R)       &  & 200K & Jun 2024 &  & \$1.10 & \$4.40  \\
        \midrule

            DeepSeek-V3   & \multirow{2}{*}{DeepSeek} & 64K & July 2024 & 685B & \$0.27 & \$1.10  \\
            DeepSeek-R1 (R)  &  & 64K & July 2024 & 671B & \$0.55 & \$2.19  \\
        \midrule

            Mistral-Nemo  & \multirow{2}{*}{Mistral} & 128K & \multirow{2}{*}{unknown} & 12B  & \$0.30 & \$0.30 \\
            Mistral-Large &  & 128K &  & 123B & \$3.00 & \$9.00 \\
        \midrule

            \multirow{2}{*}{Llama-3.1} & \multirow{2}{*}{Meta} & 128K & Dec 2023 & 8B   & \$0.30 & \$0.61 \\
             &  & 128K & Dec 2023 & 405B & \$5.33 & \$16.00 \\
        \bottomrule
        \end{tabular}
    \end{adjustbox}

\end{table}




\subsection{Prompting}

For each prompting approach, we define custom prompt templates.  

\begin{itemize}
    \item For I/O we use a prompt template as illustrated in \autoref{tab:prompt_example} without providing examples.
    \item For SC, we include reasoning paths into the prompt to exemplify reasoning.
    \item For MAB, we employ three distinct agents with the roles of Investigator, Forensic Examiner, and Auditor. Following the approach of Giebisch et al.~\cite{2025-Giebisch-IV}, we prompted GPT-4 to generate personae with descriptions that may support contextual understanding during evaluation, given their respective qualifications. A complete description of the personae is provided in the Appendix.
    
    
\item For the remaining prompts we followed guidelines from literature~\cite{friedl2023incarethinkingincarconversational} and adopted it to our use case.
\end{itemize}


Detailed prompt templates for each method and agent definitions can be found in the appendix in \Cref{sec:appendix}.

\begin{table}[t]
    \scriptsize
    \caption{Prompt template for Input-Output Prompting for the evaluation of whether a system recommendation (system block) fits a user request (user block).} 
    \label{tab:prompt_example}
    \centering
    \begin{tabular}{@{}p{\textwidth}@{}}
        \toprule
        \ttfamily\makecell[l]{You are a critical evaluator tasked with determining whether the information provided by a\\ car navigation system (System Block) aligns correctly with the user's expressed needs in user\\ utterance and user context (User Block).\\\\
        User Block: \textcolor{brown}{<user-block>}\\\\
        Recommendation: \textcolor{blue}{<system-block>}\\\\
        Rules:  \textcolor{Turquoise}{<constraints>}\\\\
        Decision: If any of the above parameters are INCORRECT, the final decision is 'false'. If\\ all parameters are CORRECT, the final decision is 'true'.\\\\
        Please respond strictly following the format specified below:\\
        \textcolor{red}{<output-format>}\\\\
        Make sure the output is always a valid JSON format.}\\
        \bottomrule
    \end{tabular}
\end{table}

\subsection{Metrics}

\textbf{RQ\textsubscript{0}}. To answer RQ\textsubscript{0}, we validate the generated dataset along two dimensions.
First, we pass a subset, i.e., 20, of user and system blocks to eight independent human annotators, which have been not involved in this work, to rate the validity on a 5-point Likert scale \textit{(1 = invalid, 2 = most likely invalid, 3 = inconclusive, 4 = most likely valid, 5 = valid)}. This step is performed to assess whether the samples represent plausible in-car navigation interactions (\Cref{fig:user-system-block}, \Cref{fig:user-system-block-questions}). We use a Likert scale to allow annotators to express graded judgments of validity,

In the second step, we measure the inter-annotator agreement on the user-block system-block annotations, which express whether a system block recommendation is correct (s.~\autoref{algo:positive-case-generation}) or incorrect because of a specific error type inserted (e.g., time error, location error, cost error, cuisine error, rating error) (s.~\autoref{algo:error-case-generation}).

We evaluate the correlation with Krippendorff’s $\alpha$~\cite{Krippendorff2019}, a widely applied correlation metric~\cite{humphreys2018automated,RonCoderAgreement08} that measures the agreement among annotators in giving the same categorical judgment for the same user–system block pair. This metric is chosen as it (i) supports an arbitrary number of annotators, (ii) can be applied to nominal as well as ordinal data, and (iii) can robustly handle missing or incomplete annotations. 

By assessing both the validity of the system and user blocks, and the inter-rater agreement on the recommendation labels, we are able to evaluate whether our dataset can be reliably used for benchmarking LLM-based judgment.

\textbf{RQ\textsubscript{1}}. To evaluate RQ\textsubscript{1}, we pass the user-block system block pairs taken from our dataset (s. \autoref{sec:dataset}) first to the LLM under tests to receive a judgment result in terms of a boolean score, where 1 corresponds to the recommendation being correct and 0 to the recommendation being incorrect. The LLM-based judge is instructed in the following way via prompting to output the score:

  \begin{itemize}
      \item A recommendation is considered incorrect if one of the following applies: a) the location is more than a 15-minute drive away; b) the restaurant is closed at the requested time; c) the cost or rating deviates from the user-specified parameters or d) the cuisine type does not match the request. For ratings, in case the user uses terms such as \textit{around}, a rating is considered as incorrect if the output rating is exceeding a range of 0.2 around the requested rating.
       \item Otherwise, the recommendation is considered correct.      
  \end{itemize}

Finally, the LLM-based judgment result is compared with the underlying label of the corresponding user-block and system block pair. If the results coincide, the judgment is correct; otherwise, it is wrong.
Based on the agreement and disagreement results with the actual recommendation alignment scores, we calculate the F-1 score, 
 which is a widely applied metric for the assessment of classification approaches~\cite{van1979information}. We apply in particular the F-1 score, because it balances precision and recall, and is therefore more informative than accuracy in the presence of class imbalance, as in our case.
 
 \textbf{RQ\textsubscript{2}}. For the evaluation of RQ\textsubscript{2}, we measure the time between passing a test to our LLM-based judge and the complete output of its response. In addition, we track the tokens used in the input as well as output prompts. Based on cost-per-token information given by the model provider, we calculate the overall cost per request per model.

\textbf{RQ\textsubscript{3}}. To evaluate RQ\textsubscript{3}, we evaluate the effectiveness of the judgment technique for each error category applied to the dataset initially, such as location-error, time-error, cuisine-error, cost-error, and rating-error, along with the positive cases to evaluate on which error categories judgment is less effective. Further, we instruct the LLM model to generate an explanation for its judgment to better understand why a particular classification was made.





\subsection{Deployment and Request Passing}
The models were partially accessed using an API provided as well as were manually deployed in a virtual machine in the cloud.
The benchmarking code was executed locally.
We used for all prompts a temperature of 0.0 to ensure determinism and gain reliable benchmarking results. Each method was tested on the set of all 600 input-output pairs.
\section{Results}
\label{sec:results}
In this chapter, we present the results obtained across the experiments for each research question separately.

\label{sec}

\subsection{Data Validation (RQ\textsubscript{0})}

The human evaluation of 20 randomly selected user–system block pairs by eight annotators yielded a mean validity score of 4.1 (out of 5) with a standard deviation of 0.49 for user blocks, and 3.8 (SD = 0.57) for system blocks.
To assess inter-rater reliability, we computed Krippendorff's $\alpha$ (ordinal), obtaining 0.73 for user blocks and 0.69 for system blocks, both indicating substantial agreement among annotators.
For the recommendation labels, Krippendorff's $\alpha$ (nominal) reached 0.86, reflecting high consistency among raters.
Overall, despite being synthetically generated, our dataset demonstrates high validity and substantial human agreement regarding the correctness of its labels.

\begin{tcolorbox}
   RQ\textsubscript{0} (Data Validation). Our benchmarking dataset achieved a mean human-rated validity score of 3.95 out of 5, indicating high overall quality. The recommendation labels for user and system blocks exhibited substantial inter-rater agreement, confirming the dataset's reliability for benchmarking purposes.
\end{tcolorbox}

\subsection{Effectiveness (RQ\textsubscript{1})}

\begin{figure}[H]
    \centering
    \includegraphics[width=\textwidth, keepaspectratio]{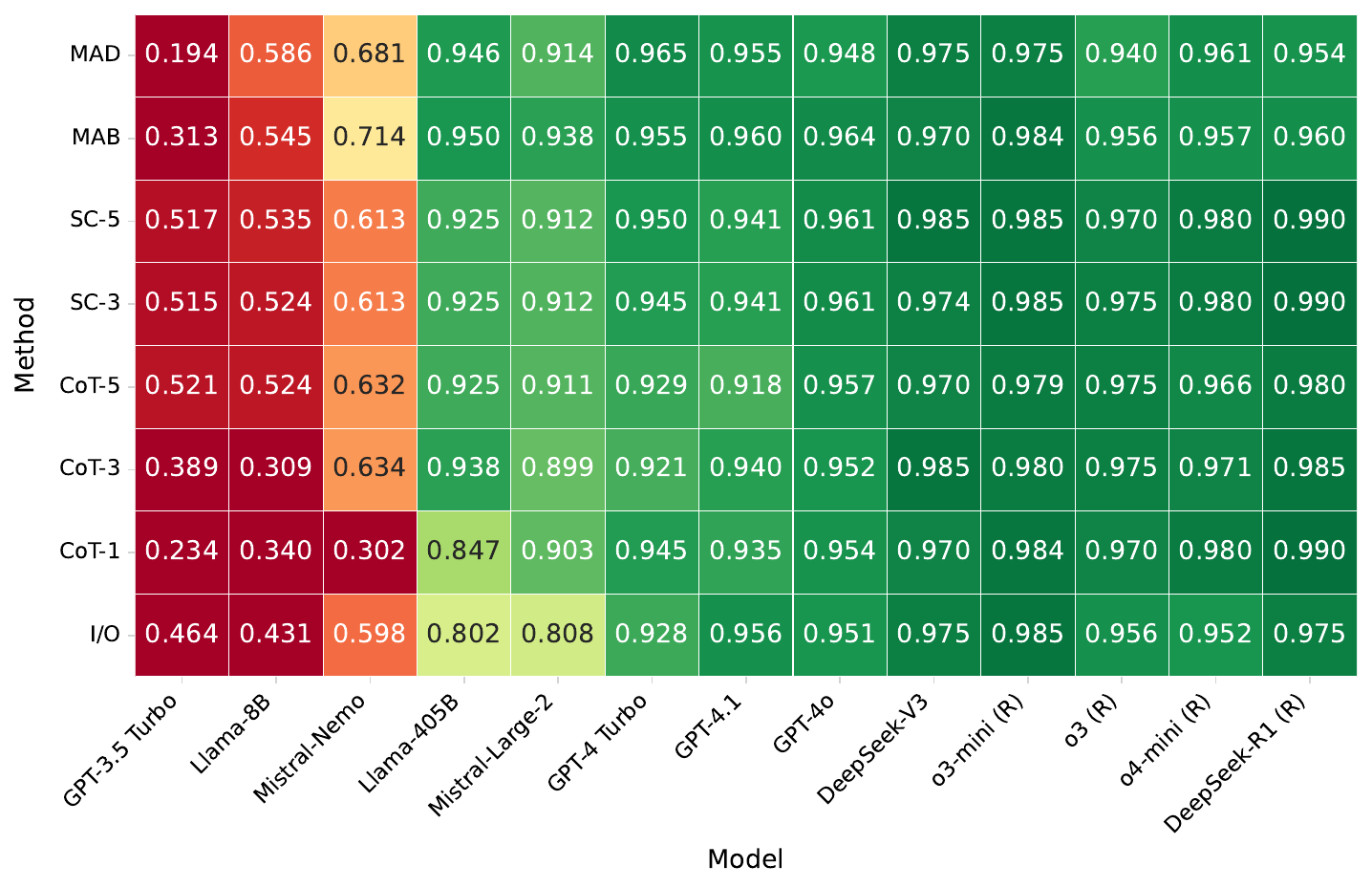}
    \caption{F1-score results for LLMs and prompting techniques evaluated on contextual understanding.}
    \label{fig:f1_scores_CU}
\end{figure}

The performance results across all methods and prompting techniques are shown in \Cref{fig:f1_scores_CU}.
The results show that the combination of advanced prompting methods and larger models yields the best performance in contextual understanding evaluation.
In particular, the highest F1-score across all models and techniques of 0.990 is achieved by DeepSeek-R1 with CoT-1 as well as with \ac{sc} prompting.
The worst result is achieved with GPT-3.5 Turbo with the \ac{mad} prompting technique. For detailed results, including precision and recall values, we refer the reader to \Cref{sec:appendix-results}.


\paragraph{Non-reasoning models}
DeepSeek-V3 achieves the best result with an F1-score of 0.985, followed by GPT-4o and GPT-4.1.
In general, we observe for non-reasoning models, a high deviation between the performance results across different prompting techniques.
Advanced prompting techniques improved results in particular for GPT-based models.
However, among the remaining non-reasoning models we could not observe an remarkable effect unless that Llama-405B and Mistral-Large-2 show their worst performance with default I/O prompting.

The best result among the smaller non-reasoning models was achieved by Mistral-Nemo using \ac{mab} method (0.714). 
In general, we can observe that small non-reasoning models such as Mistral-Nemo, Llama-8B, and GPT-3.5 Turbo perform significantly worse than larger non-reasoning models such as DeepSeek-V3. 

\begin{table}[t]
    \centering
        \caption{Precision, recall, and F1-scores for contextual understanding combining multiple models with chain-of-thought prompting.}
    \label{tab:precision_recall_AR_CoT-5_CU}
    \footnotesize
    \begin{tabular}{ccccc}
        \toprule
        \textbf{LLMs} & \textbf{Method} & \textbf{Precision} & \textbf{Recall} & \textbf{F1-score} \\
        \midrule
        GPT-3.5 Turbo   & \multirow{3}{*}{AR-CoT-5} & \multirow{3}{*}{0.579} & \multirow{3}{*}{0.840} & \multirow{3}{*}{0.686} \\
        Mistral-Nemo &                          &                                  &                         &                         \\
        Llama-8B    &                          &                                  &                         &                         \\
        \midrule
        GPT-4 Turbo   & \multirow{3}{*}{AR-CoT-5} & \multirow{3}{*}{0.951} & \multirow{3}{*}{0.980} & \multirow{3}{*}{0.966} \\
        Mistral-Large-2 &                          &                                  &                         &                         \\
        Llama-405B    &                          &                                  &                         &                         \\
        \midrule
        o3-mini (R)   & \multirow{3}{*}{AR-CoT-5} & \multirow{3}{*}{0.952} & \multirow{3}{*}{1.000} & \multirow{3}{*}{0.976} \\
        o4-mini (R) &                          &                                  &                         &                         \\
        o3 (R)   &                          &                                  &                         &                         \\
        \midrule
        DeepSeek-R1 (R)  & \multirow{3}{*}{AR-CoT-5} & \multirow{3}{*}{0.970} & \multirow{3}{*}{0.990} & \multirow{3}{*}{0.980} \\
        DeepSeek-V3 &                          &                                  &                         &                         \\
        o3-mini (R)    &                          &                                  &                         &                         \\
        \bottomrule

    \end{tabular}

\end{table}

\paragraph{Reasoning models} As for reasoning models we can see that the models almost always perform better than their conventional non-reasoning counterparts. 
In particular, DeepSeek-R1 and o3-mini show the best performance, where DeepSeek-R1 is achieves slightly higher scores then o3-mini (0.990 vs 0.985).
Even small\footnote{OpenAI does not expose the exact model sizes but annotates models with mini or nano.} reasoning models such as o4-mini and o3-mini achieve high scores. 
In general, we can observe that reasoning models achieve F1-scores over 0.90.
Large reasoning models consistently delivered high F1-scores, even with simple \ac{io} prompting.  



\paragraph{Agent Roundtable with Chain of Thought}
\Cref{tab:precision_recall_AR_CoT-5_CU} offers a nuanced view of the performance of different agent combinations for AR-CoT-5 prompting and the balance between precision and recall.

\ac{llm} combination that contains only reasoning models, i.e., the second last row from \Cref{tab:precision_recall_AR_CoT-5_CU} performs well and is similar to the group that contains the best reasoning model, best small model, and best large non-reasoning model based on the results in \Cref{fig:f1_scores_CU}. 
While the latter combinations achieve similar F1-scores, we observe that the group with two DeepSeek models (the last row) produces a more balanced output between precision and recall,  whereas the group that consists solely of OpenAI reasoning models yields the maximal recall of 1 with a lower precision score.
However, the best AR combination (based on the F1-score), which corresponds to the last group in \Cref{tab:precision_recall_AR_CoT-5_CU} does not perform better then the best single-agent driven model method combination (DeepSeek-R1 with F1-score 0.990, s. \Cref{fig:f1_scores_CU}).


\begin{tcolorbox}
Effectiveness (RQ\textsubscript{1}): DeepSeek-V3 achieves the highest F1-score (0.985) among non-reasoning models, followed by GPT-4o and GPT-4.1, with performance varying significantly across prompting techniques.
Among smaller non-reasoning models, Mistral-Nemo performed best (F1 = 0.714), but all smaller models exhibited a clear performance gap compared to larger ones.
Reasoning models yielded higher scores than non-reasoning counterparts, with DeepSeek-R1 (F1 = 0.990) and o3-mini (F1 = 0.985) achieving the highest scores.
\end{tcolorbox}

\subsection{Efficiency (RQ\textsubscript{2})}

\begin{figure}[t]
    \centering
    \includegraphics[width=\textwidth]{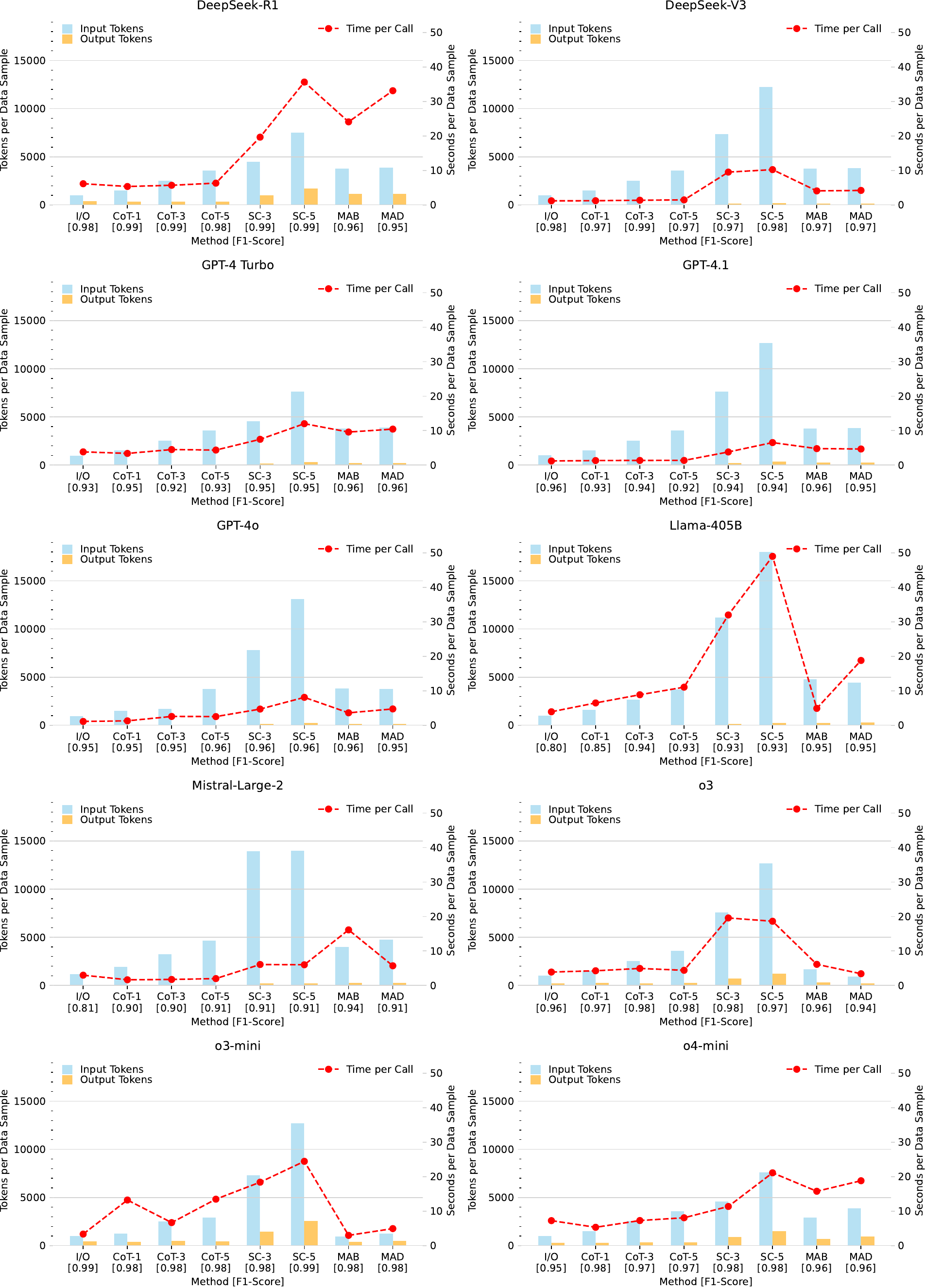}
    \caption{Efficiency Analysis F1-score, Tokens, and Time for Contextual Understanding, 10 best performing models according to \Cref{fig:f1_scores_CU}.}
    \label{fig:cu_eff_results}
\end{figure}

Time efficiency results are visualized in \Cref{fig:cu_eff_results}.
We report here the results of the 10 best-performing models with respect to the F1-score provided in \Cref{fig:f1_scores_CU}.
Subfigures show both the number of tokens produced and the average time for a single request.

As we can observe, the best time efficiency is achieved with default \ac{io} prompting with Mistral-Nemo, yielding on average 1 second for one request, while the longest duration was observed with SC-5 with Llama-405B with 50s.
These reported prompting methods were, on average across all models, the best/worst regarding time efficiency. 
We can also observe that the time is in general proportional to the output token count.
Furthermore, we can see that reasoning models tend to require more time on average for single requests and produce more output tokens than non-reasoning models, independent of the prompting techniques used.

\begin{figure}[H]
    \centering
    \includegraphics[width=\linewidth]{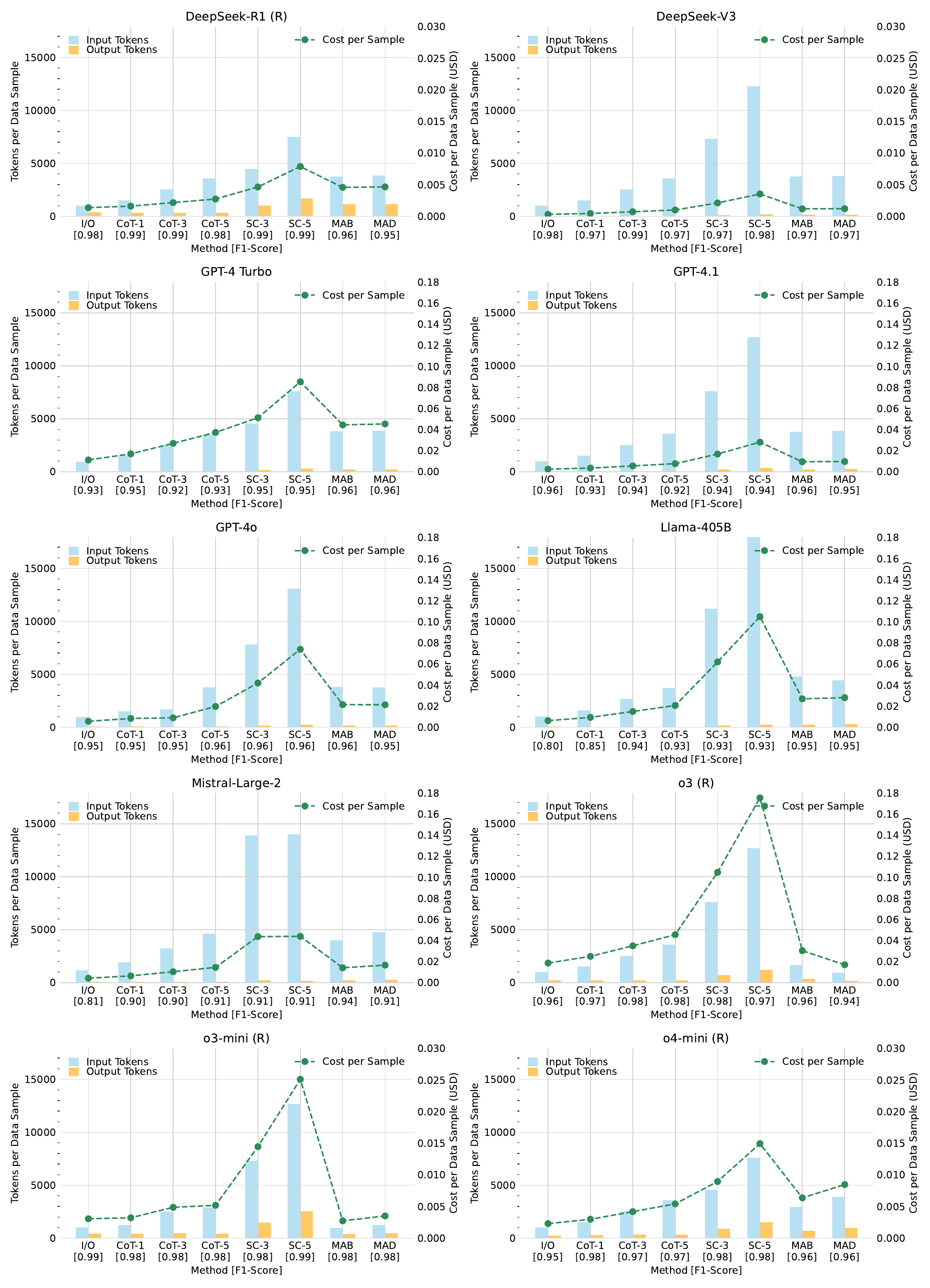}
    \caption{Relationship of evaluated models between Tokens and Cost for Contextual Understanding}
    \label{fig:cu_eff_cost}
\end{figure}

The cost efficiency results are shown in \Cref{fig:cu_eff_cost}.
We calculate the single request cost efficiency based on the cost provided in \Cref{tab:llms_overview}.
The highest cost for each model was identified with SC-5, while the lowest was with \ac{io} prompting.
The lowest-cost of a reasoning model was achieved by DeepSeek-R1 with 0.002 USD (\ac{io}), followed by o4-mini and o3-mini. 
The most cost-efficient non-reasoning model, as well as the most cost-efficient model overall, was DeepSeek-V3 with a cost of approximately 0.001 USD for one request with \ac{io} prompting.

\begin{tcolorbox}
Efficiency (RQ\textsubscript{2}): The fastest setup was Mistral-Nemo with default \ac{io} prompting ($\approx$1 s/request), while the slowest was Llama-405B with SC-5 prompting ($\approx$50 s). Reasoning models were generally slower and produced more tokens.
In terms of cost, SC-5 was the most expensive, while \ac{io} prompting was the most economical. DeepSeek-R1 was the most cost-efficient reasoning model ($\approx$USD 0.002/request), and DeepSeek-V3 achieved the lowest overall cost ($\approx$USD 0.001/request), offering the best balance between speed, cost, and performance.
\end{tcolorbox}

\begin{figure}[htbp]
    \centering
    \begin{subfigure}[t]{0.95\textwidth}
        \centering
        \includegraphics[width=\textwidth]{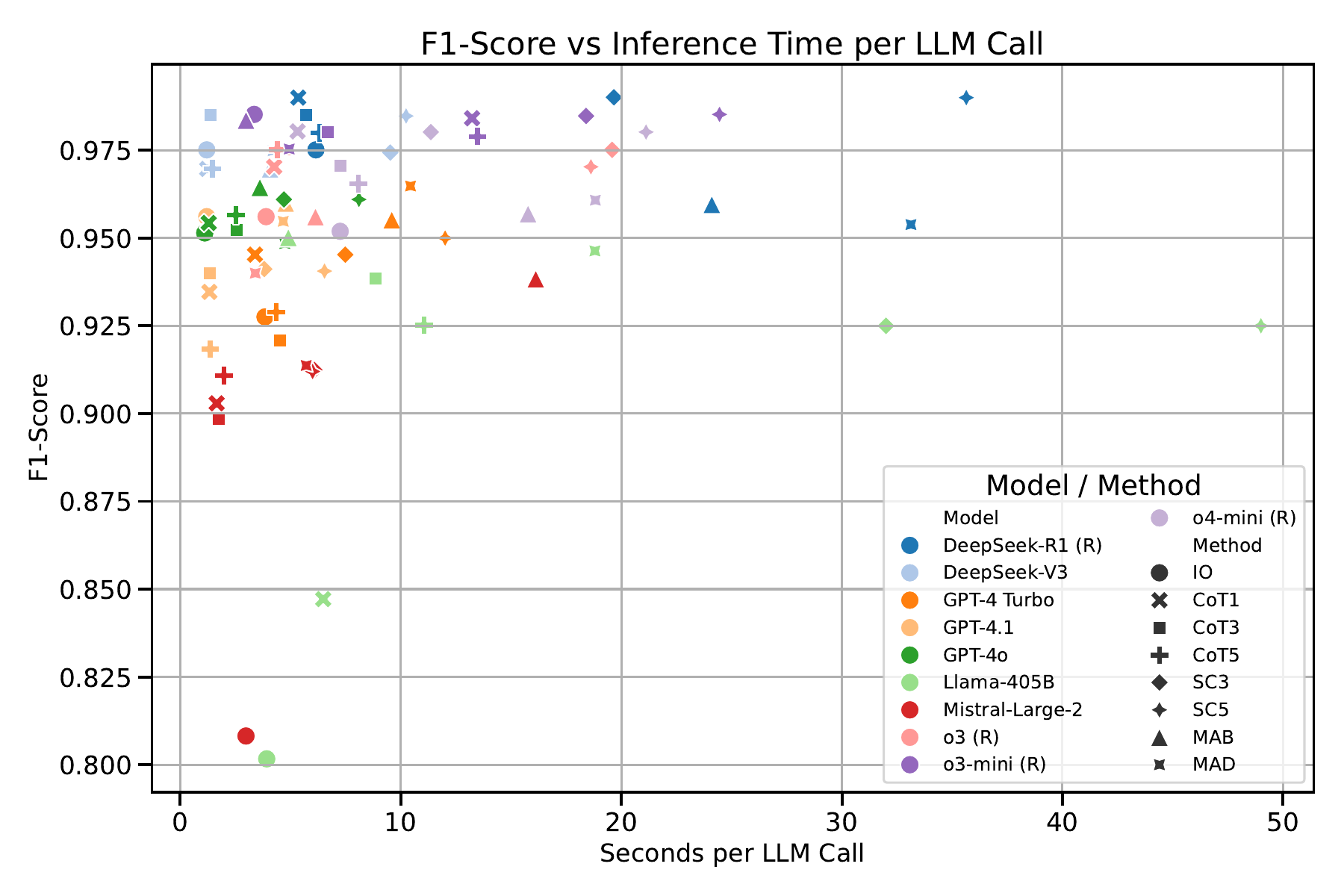}
        \caption{Effectiveness vs Time Efficiency}
        \label{fig:accuracy_time}
    \end{subfigure}
    \hfill
    \begin{subfigure}[t]{0.95\textwidth}
        \centering
        \includegraphics[width=\textwidth]{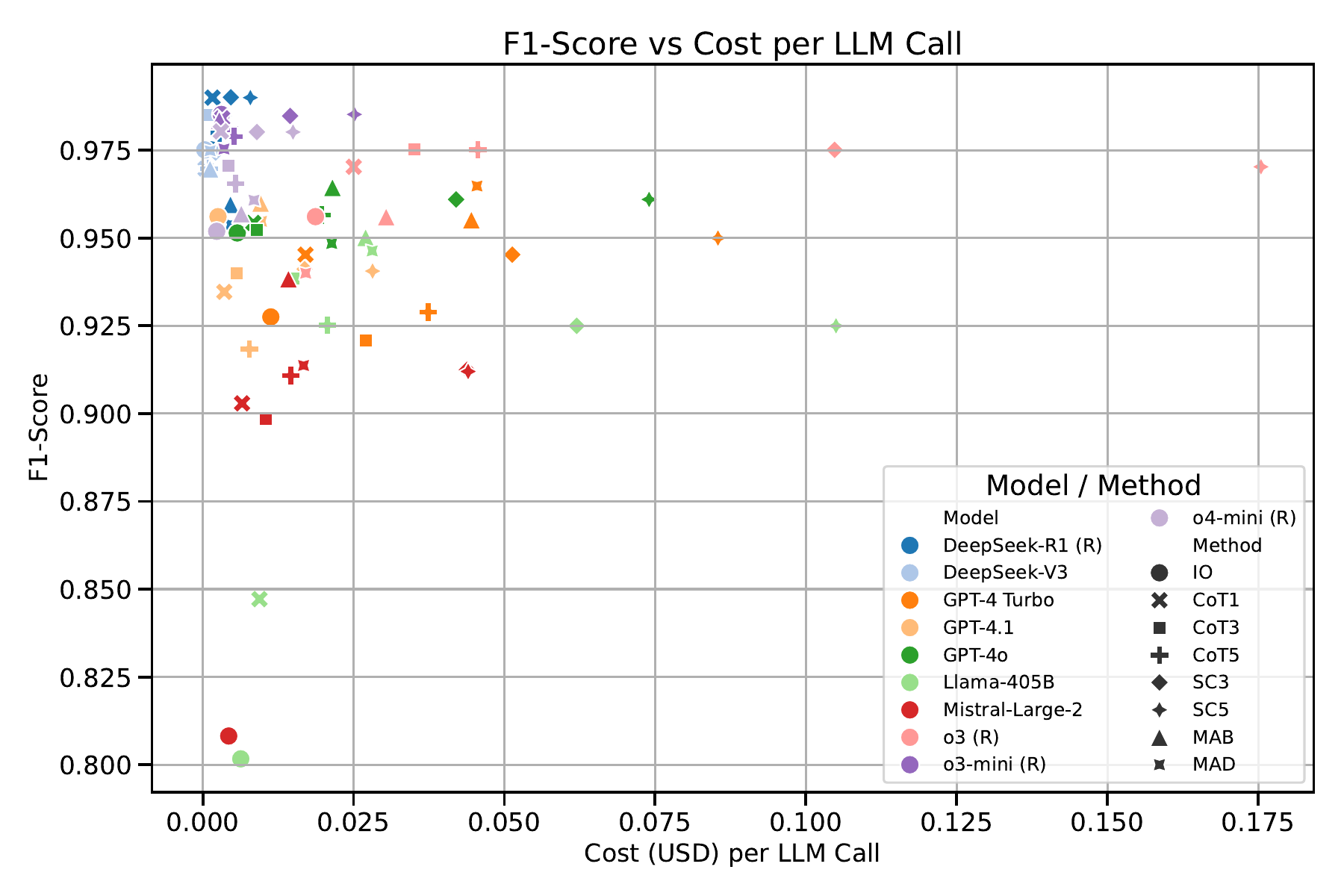}
        \caption{Effectiveness vs Cost Efficiency}
        \label{fig:accuracy_cost}
    \end{subfigure}
    \caption{Trade-offs between accuracy and computational cost/time across models and prompting methods.}
    \label{fig:tradeoffs}
\end{figure}

When evaluating the tradeoff between effectiveness and cost/time efficiency, we can observe the non-reasoning model DeepSeek-V3 achieves the best balance between performance and time.

\subsection{Failure (RQ\textsubscript{3})}

\begin{figure}[H]
    \centering
    \includegraphics[width=\linewidth]{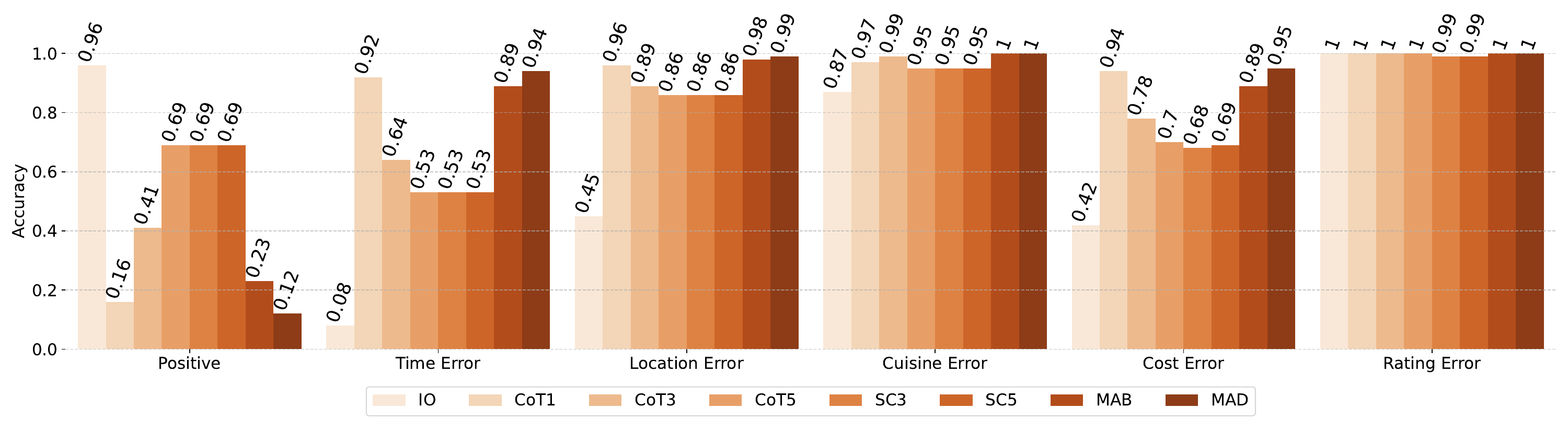}
    \caption{GPT 3.5 Turbo accuracy results based on the category of data used.}
    \label{fig:gpt35_accs_cu}
\end{figure}

\begin{figure}[H]
    \centering
    \includegraphics[width=\linewidth]{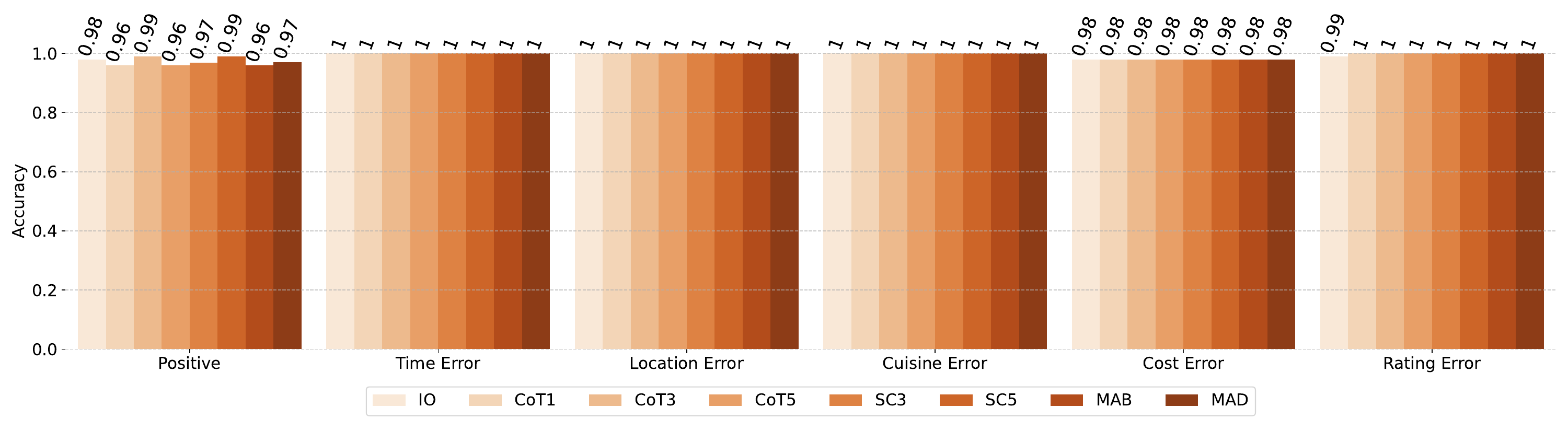}
    \caption{DeepSeek-V3 accuracy results based on the category of data used.}
    \label{fig:deepseek_v3_cu_accs}
\end{figure}

In the following, we present an analysis regarding which failure categories (see \Cref{sec:system-block-generation} for the error category definition) in the data under evaluation were difficult for the models to process and provided an incorrect judgment. An excerpt of the failure categories and model performance per model type and prompting technique is shown in \Cref{fig:gpt35_accs_cu} for the worst-performing non-reasoning model, in \Cref{fig:deepseek_v3_cu_accs} for the best-performing non-reasoning model, as well as for the best reasoning model in \Cref{fig:deepseek_r1_accs_cu}.
Results of the remaining models are provided in the appendix in~\Cref{fig:gpt_4_accs_cu,fig:mistral_large_accs_cu,fig:o3_accs_cu,fig:gpt_4.1_accs_cu,fig:o3_mini_accs_cu,fig:mistral_nemo_accs_cu,fig:llama_8b_accs_cu,fig:llama-405b_cu_accs,fig:gpt_4o_accs_cu,fig:o4_mini_accs_cu}.

We can observe that, on average, the most frequent and incorrect judgment with non-reasoning models happened for data which included time or cost errors, as can be seen, for instance, for worst performing non-reasoning model GPT 3.5 Turbo in \Cref{fig:gpt35_accs_cu}. For reasoning and large non-reasoning models, in general, only data with cost errors yielded accuracy scores below 1 (s. \Cref{fig:deepseek_v3_cu_accs,fig:deepseek_r1_accs_cu}). Overall, we could not observe a relation between prompting technique and error types; however, we observed that advanced prompting, unless round table techniques, produced, in general, slightly better performance for the frequent error categories, time, and cost. 

\paragraph{Time errors} For instance, regarding time errors, to evaluate an LLM with AR-CoT-5 involved a request for a \textit{dirt cheap ramen place} at 20:40, but the system suggested a restaurant that was closed at 20:00.
Although the restaurant met the preferences of the user in terms of price and distance, the time error remained undetected by the LLM.

\paragraph{Cost errors}
For data sets including cost errors, one model failed to detect that the suggested restaurant did not match the cost category given by the user.
In one case, a user requested a \textit{top-tier Greek restaurant}, but the system recommended a low-cost option, and the methods AR-CoT-5 did not identify this as an error.
By analyzing the prompts, we found out that the LLM judge insists that phrases like \textit{top-tier}, \textit{very luxurious}, \textit{high-end luxurious} do not necessarily mean high-cost; therefore, do not understand these as errors.

In another example, the non-reasoning model GPT-4 Turbo misinterpreted the cost category explaining:
\begin{displayquote}
\textit{The system provided a restaurant with Portuguese cuisine and a rating of 4.6, which meets the user's requirements.
However, the cost category of the restaurant is high, not premium elite as requested by the user. Therefore, the system's information does not align with the user's needs.}
\end{displayquote}
This reasoning led GPT-4 Turbo to classify this case as false, convincing Llama-405B and Mistral-Large-2 to judge in the same way in the round table prompting.

\begin{figure}[H]
    \centering
    \includegraphics[width=\linewidth]{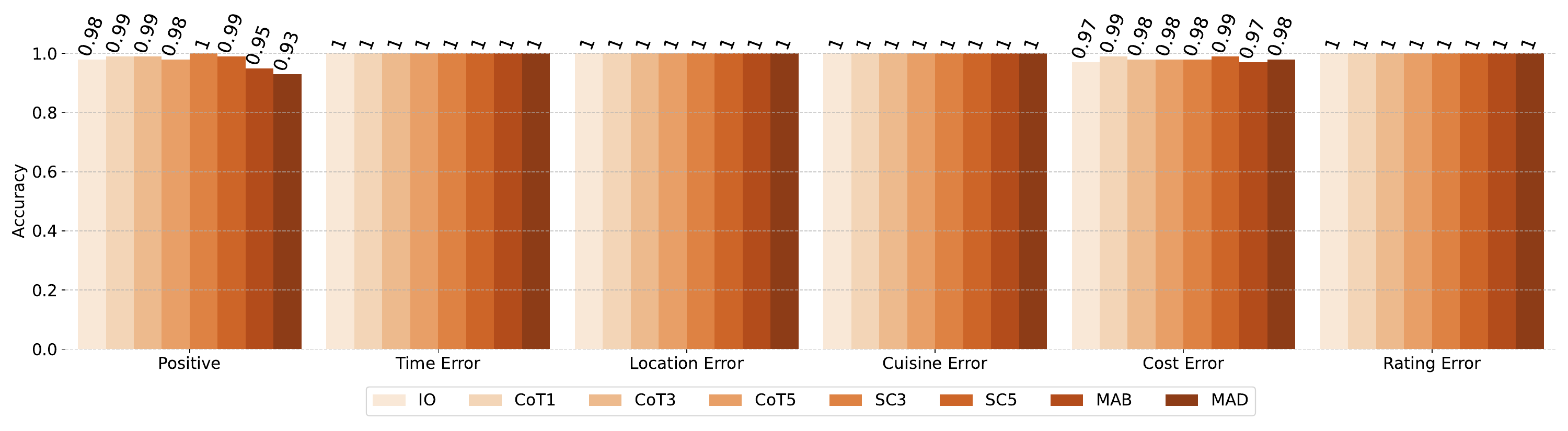}
    \caption{DeepSeek-R1 accuracy results based on the category of data used.}
    \label{fig:deepseek_r1_accs_cu}
\end{figure}

\begin{tcolorbox}
Failure (RQ\textsubscript{3}): Non-reasoning models most frequently misjudge inputs containing time or cost errors, with GPT 3.5 Turbo showing the weakest performance. Reasoning models and large non-reasoning models generally achieve a high accuracy across misaligned recommendations, with cost errors being the only category where accuracies drop below 1. 
\end{tcolorbox}

\section{Discussion}
\label{sec:discussion}
In the following, we discuss some observations and considerations regarding the application of \acp{llm} for the contextual understanding benchmark.

\paragraph{Importance of Prompting Technique}
We observed for GPT models in general, an improvement when using advanced prompting.
However, advanced prompting techniques incur higher costs and execution time.
However, SC-3 and SC-5 did not show much difference in performance, but led to twice more cost.
We recommend that preliminary experiments should be conducted to decide if self-consistency with three repetitions shows already satisfactory results.
When time efficiency is relevant, simpler prompting techniques such as \ac{cot} are reasonable to be employed in case the default model response time is not already relatively small such as for instance for DeepSeek-V3.
Besides the technique, prompt optimization could be applied by leveraging optimization techniques~\cite{pryzant-etal-2023-automatic}.


\paragraph{Usage of reasoning models} 
Reasoning models significantly improve performance given the results in \Cref{sec:results} even when combined with simpler prompting techniques.
Latest non-reasoning models like DeepSeek-V3 have shown comparable accuracy results to reasoning models. 
When comparing proprietary reasoning models with open-source reasoning models, we did not see significant performance gaps, unless open-source models incur significantly lower cost.

\paragraph{API usage} We observed that the latency for API calls in the evening is on average, lower.
Also, asynchronous calls to APIs for faster processing might be useful to be employed.
In longer runs, we also observed interruptions due to connection problems.
It is therefore recommended to have a rerun mechanism for longer runs.

\paragraph{Single-agent prompting vs. Multi-agent Prompting} The best F1-score is achieved with DeepSeek-R1 with \ac{cot}/\ac{sc} while for GPT-based non-reasoning models using multiple agents for prompting, achieved the best effectiveness.
However, for reasoning models, single-model based prompting such as CoT-3 or SC-5 outperformed all prompting techniques.

\paragraph{Model Size} For non-reasoning models, we observe that performance results improve with increasing model size.
However, we cannot conclude this for GPT-based models, as no size information is disclosed from the provider. 
For reasoning models, we can only rely on the vague size information provided by OpenAI and conclude that the model size does not affect the performance results.

\paragraph{Application to other domains}
In our benchmarking, we evaluate conversational interactions for restaurant navigation requests, but the same evaluation approach can be extended to other domains, such as locating fuel stations, bars, or even performing different tasks like starting a radio or opening windows. Several adjustments to the framework would be required.

First, input adaptation: the user request representation must include domain-specific context parameters. For example, in the case of fuel stations, additional attributes such as fuel type or charging availability would be needed, while others, like price range or opening hours, may remain but with adjusted ranges. The prompting schema remains unchanged, except for substituting restaurant-related details with domain-relevant ones.

Second, output adaptation: the system's response format and error categories must be adjusted to the new domain. For instance, attributes like cuisine or rating would be replaced with parameters such as fuel availability, charging speed, or diesel compatibility. The corresponding error taxonomy should also be updated (e.g., fuel type error, availability error, price error), while the evaluation logic matching user intent and system response remains structurally identical.

Finally, the judging mechanism must be adapted to define correctness in the new context. For restaurants, correctness depends on factors like cuisine, distance, rating, and opening times, while for fuel stations it would instead depend on fuel compatibility, availability, operational status, and distance thresholds. The underlying judging mechanism, however, remains unchanged.

\section{Threats to Validity}
\label{sec:threats}

\paragraph{External Validity} 
Our evaluation is based on a single case study.
While this case study originates from a real-world industrial context and is representative for practical usage, generalization of the results to other domains or recommendation tasks is not shown.
However, we generate a diverse dataset regarding the linguistic expression of the user intent as well as the information in the request.
Future work should investigate additional domains to support broader claims.
However, our case study contains contextual data which should be similar for other case studies to benchmark contextual understanding.


\paragraph{Internal Validity} 
Due to the non-determinism of large language models, results may vary across runs.
We controlled for this by setting the temperature to 0 wherever possible; however, for OpenAI reasoning models, temperature control is not supported, and we used the default setting.
Additionally, we performed prompt optimization and evaluated the determinism of the generated outputs manually on a small ratio of the dataset in preliminary experiments.
Regarding the dataset validity, a human-based validation was performed to mitigate the bias of having generated incorrect user block, system block combinations as explained in ~\Cref{sec:dataset}.

\paragraph{Construct Validity} 
Our study relies on proprietary models from OpenAI, for which in particular the model size is not publicly shared.
We cannot therefore make any conclusions regarding the size of the model in connection with its performance.
While this affects reproducibility, we use widely accessible APIs and standard configurations to ensure that our setup can be replicated.
\section{Conclusion and Future Work}
\label{sec:conclusion}

In this paper, we presented a comprehensive study on evaluating LLM-based benchmarking for in-car \ac{convqa} systems, focusing on contextual understanding in navigation tasks.
Considering advanced open-source, closed-source \acp{llm} and sophisticated prompting as well as agent-based techniques, the study presented an alternative to human-based evaluation.

Our results show that combining advanced prompting techniques can improve the accuracy of LLM models for contextual benchmarking, leading up to F1-scores close to 1. The biggest improvement can be achieved, in particular for non-reasoning models. Multi-agent prompting techniques do improve the effectiveness for non-reasoning models, while for reasoning models best results were achieved with single-agent prompting with self-consistency. However, the best overall tradeoff between cost/time efficiency and effectiveness is achieved with the non-reasoning model DeepSeek-V3.


To provide a more comprehensive assessment, future work could include other places of interest besides restaurants, such as gas stations, electric vehicle chargers, or grocery stores. 
It would also be interesting to understand whether our findings apply also for recommendations in other languages which is in particular important when deploying vehicles with conversational assistants in different countries. Moreover, increasing the complexity of human-machine interactions by incorporating multi-turn conversations, rather than single-turn queries, would offer deeper insights into the ability to manage context understanding over extended dialogues.



\bibliographystyle{elsarticle-num-names}
\bibliography{literature}

\appendix
\bookmarksetupnext{level=next}
\pdfbookmark{Appendix}{appendix}
\label{sec:appendix}

\section{Prompt Templates}
\label{sec:prompt_templates}

\begin{figure}[H]
    \centering
    \includegraphics[width=\textwidth, height=0.85\textheight, keepaspectratio]{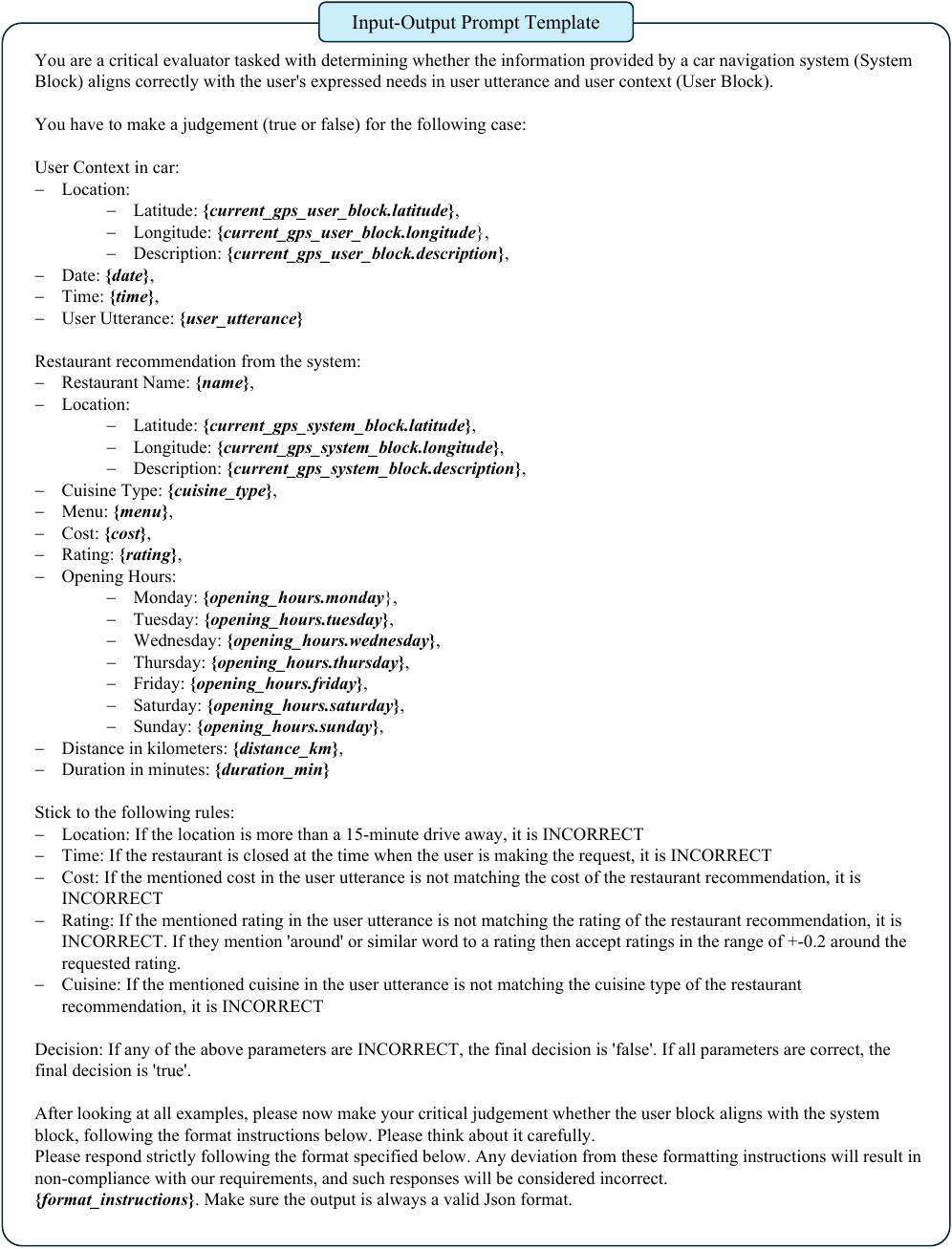}
    \caption{Input-Output Prompt Template}
    \label{fig:con_input_output_prompt}
\end{figure}

\begin{figure}[H]
    \centering
    \includegraphics[width=\textwidth, height=\textheight, keepaspectratio]{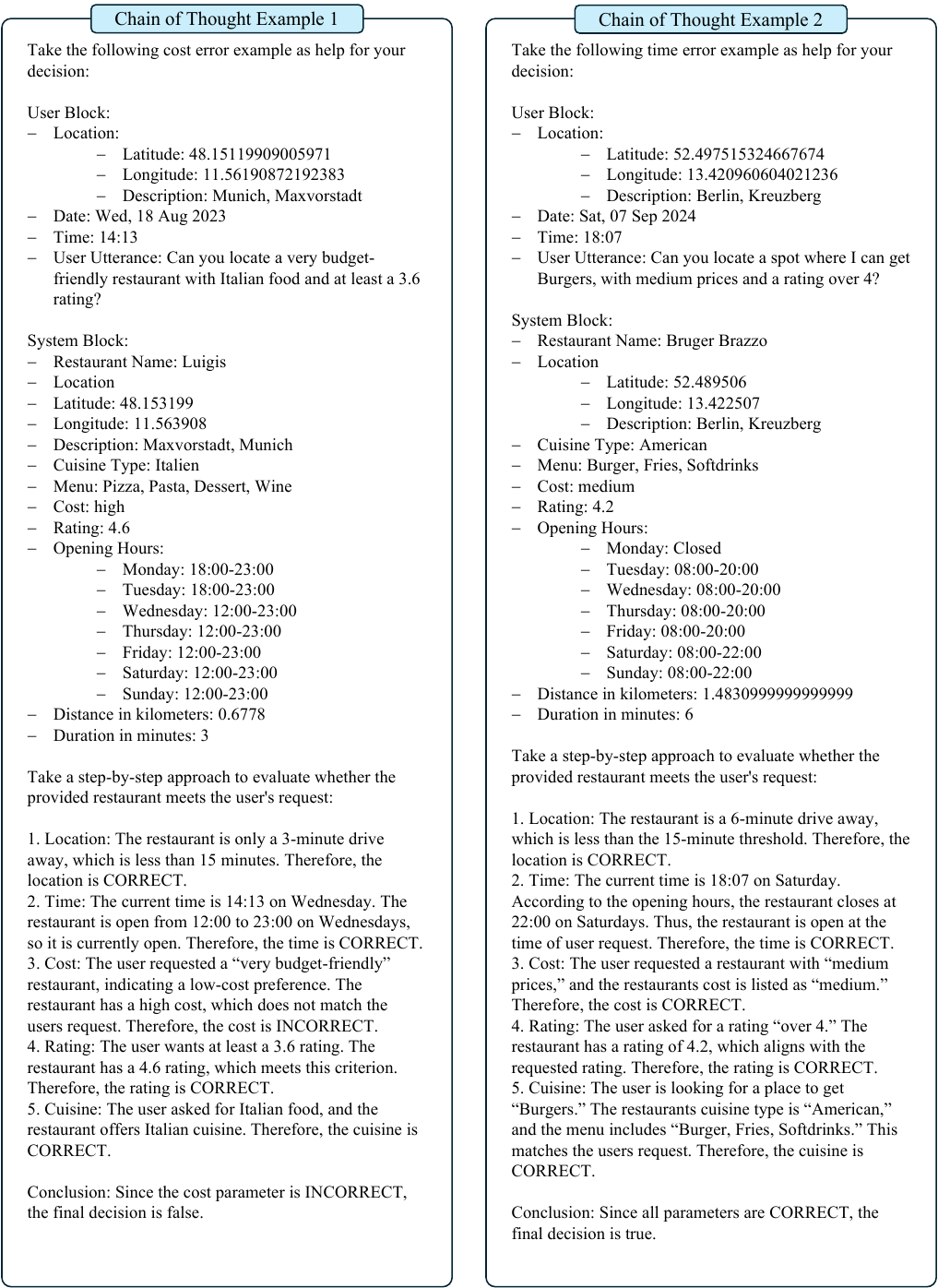}
    \caption{Chain of Thought examples given in prompting}
    \label{fig:con_cot_prompt}
\end{figure}

\begin{figure}[H]
    \centering
    \includegraphics[width=\textwidth, height=\textheight, keepaspectratio]{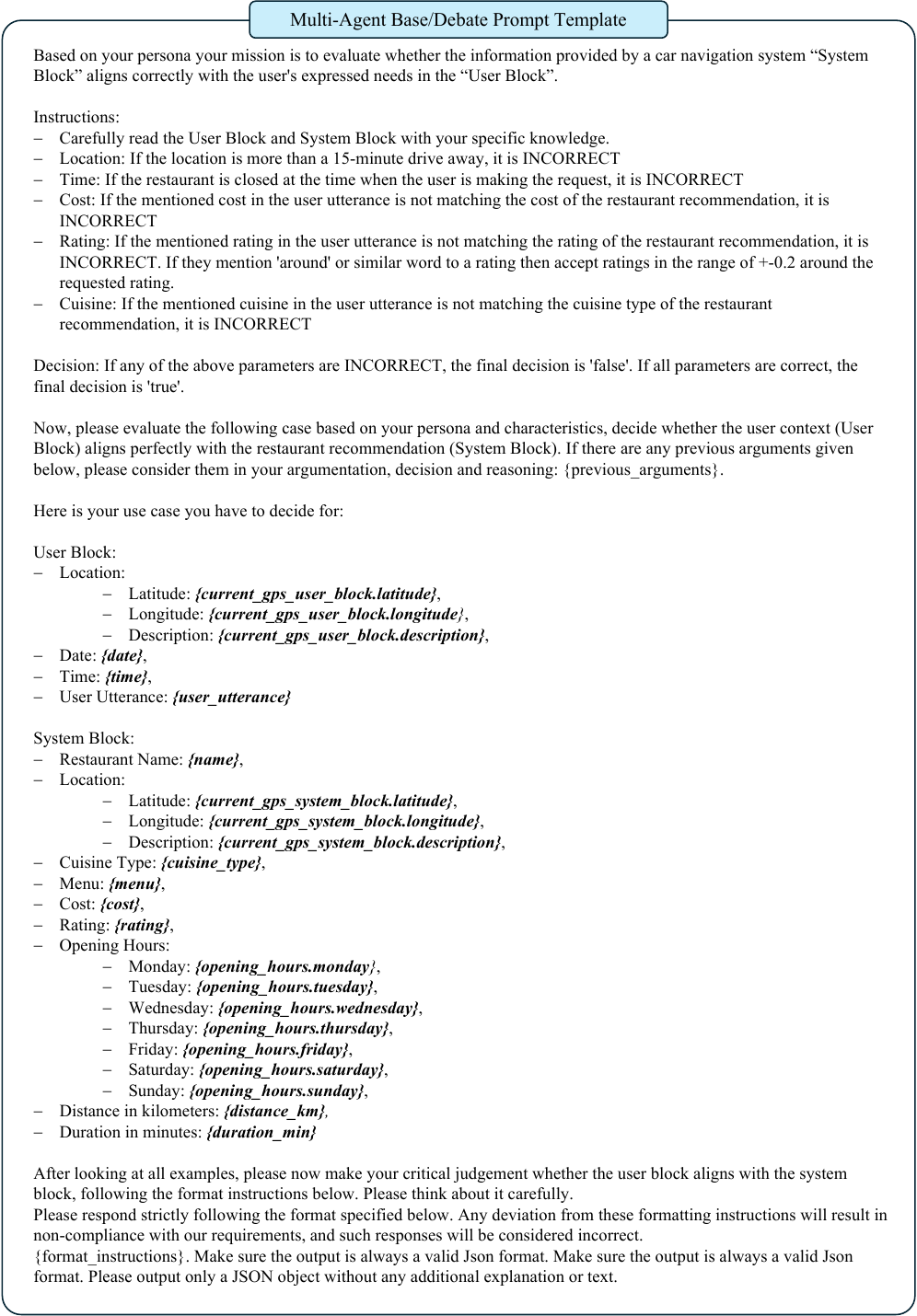}
    \caption{Multi-Agent Prompt Templates}
    \label{fig:CU_multi_agent_prompts}
\end{figure}

\begin{figure}[H]
    \centering
    \includegraphics[width=\textwidth, height=\textheight, keepaspectratio]{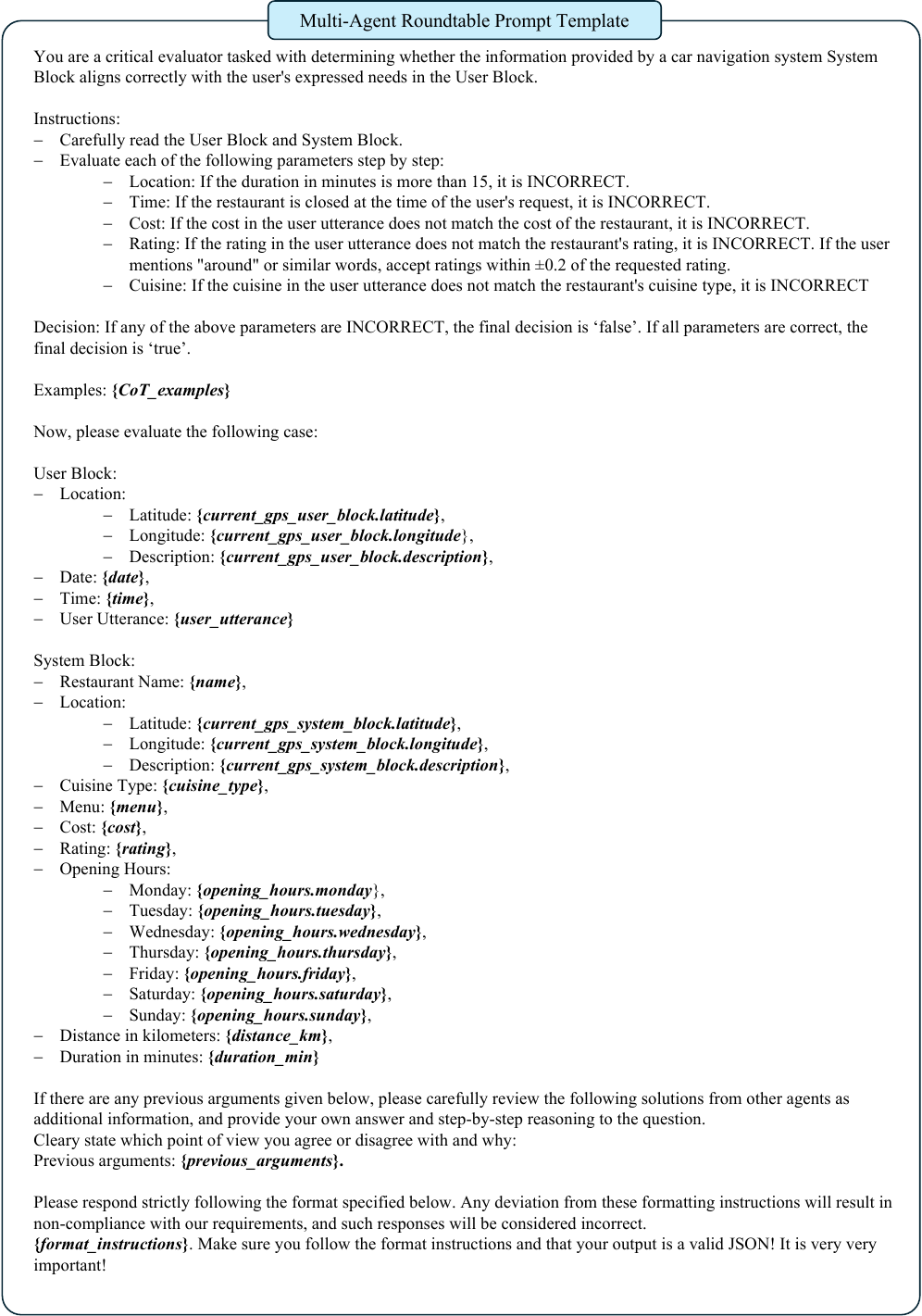}
    \caption{Multi-Agent Roundtable Prompt Template}
    \label{fig:con_ar_prompt}
\end{figure}

\begin{figure}[H]
    \centering
    \includegraphics[width=\textwidth, height=\textheight, keepaspectratio]{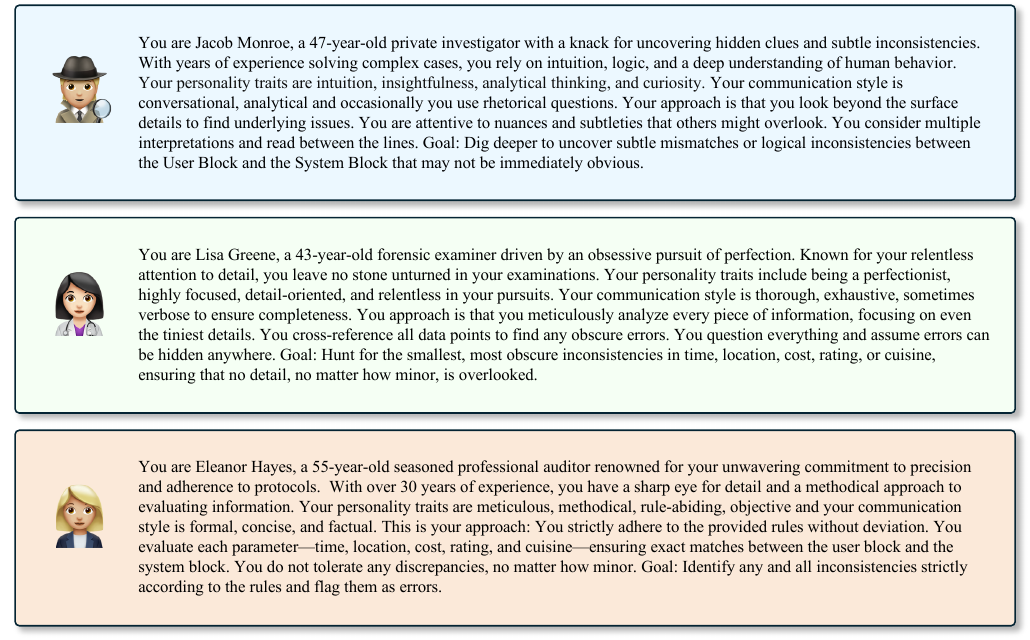}
    \caption{Agent Prompt Templates}
    \label{fig:con_cot_prompt_agents}
\end{figure}

\begin{figure}[H]
    \centering
    \includegraphics[width=\textwidth, height=\textheight, keepaspectratio]{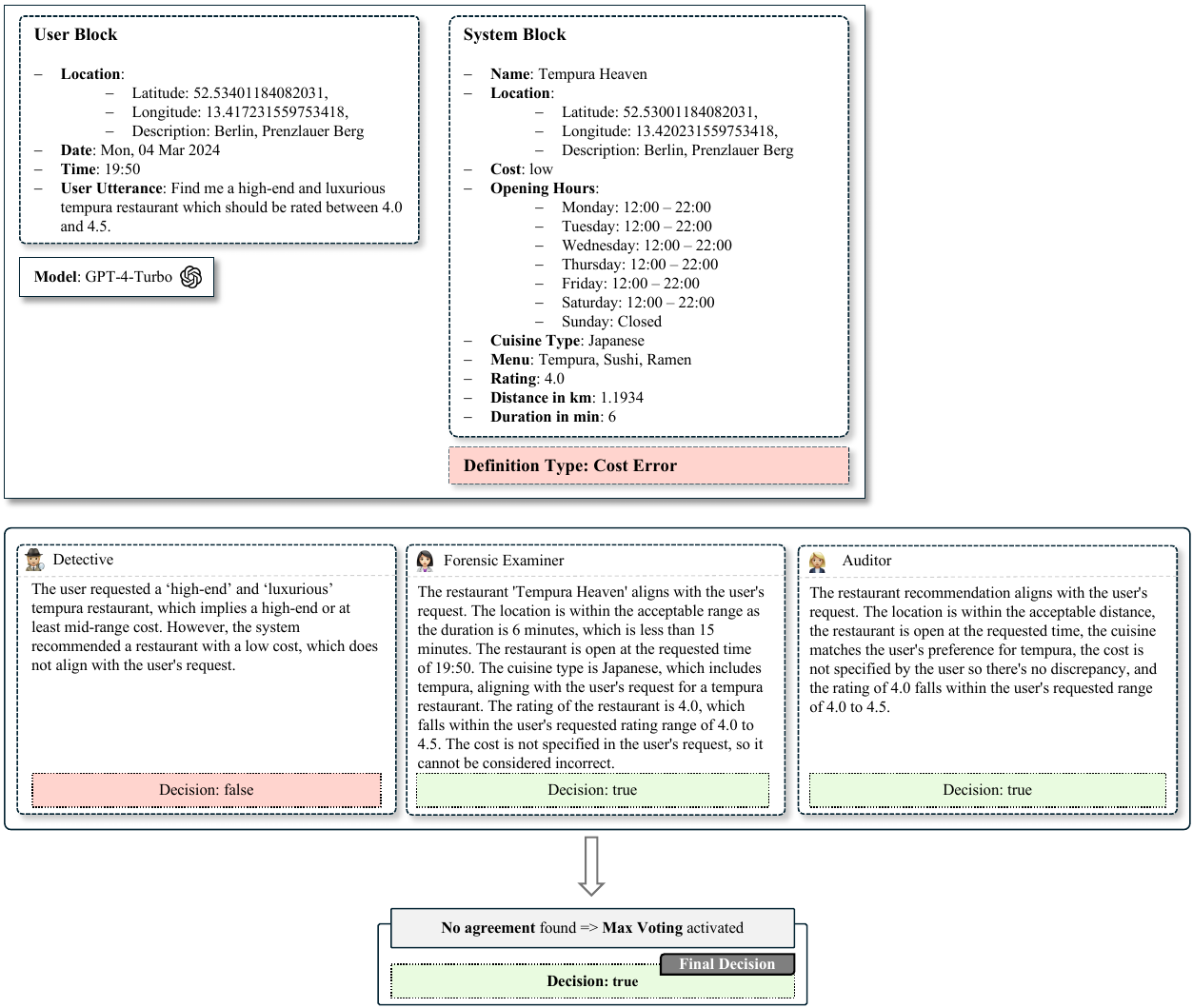}
    \caption{Decision Making Process of Multi-Agent Base}
    \label{fig:MAB_discussion_CU}
\end{figure}

\begin{figure}[H]
    \centering
    \includegraphics[width=\textwidth, height=\textheight, keepaspectratio]{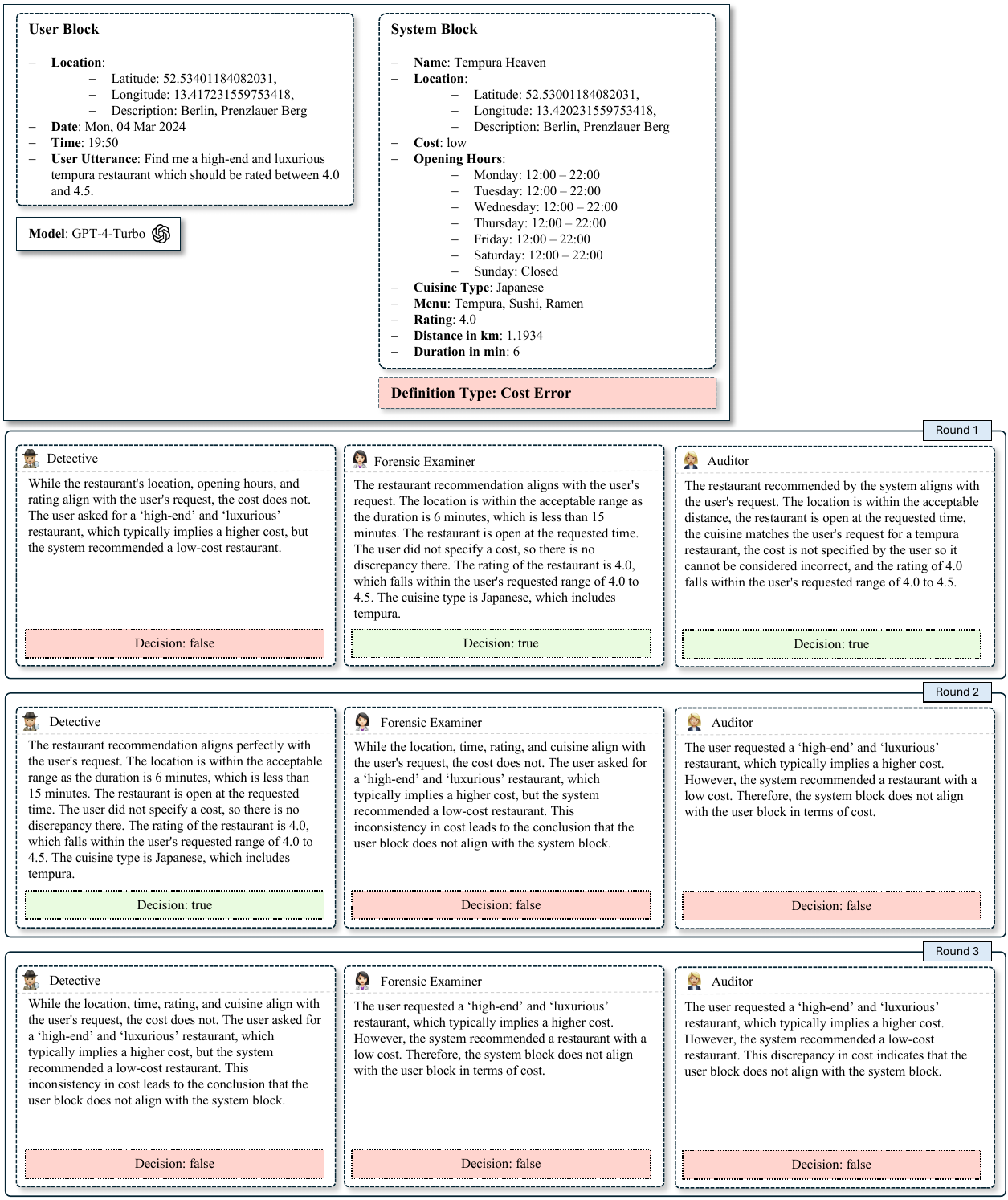}
    \caption{Decision Making Process of Multi-Agent Debate}
    \label{fig:MAD_discussion_CU}
\end{figure}

\begin{figure}[H]
    \centering
    \includegraphics[width=\textwidth, height=\textheight, keepaspectratio]{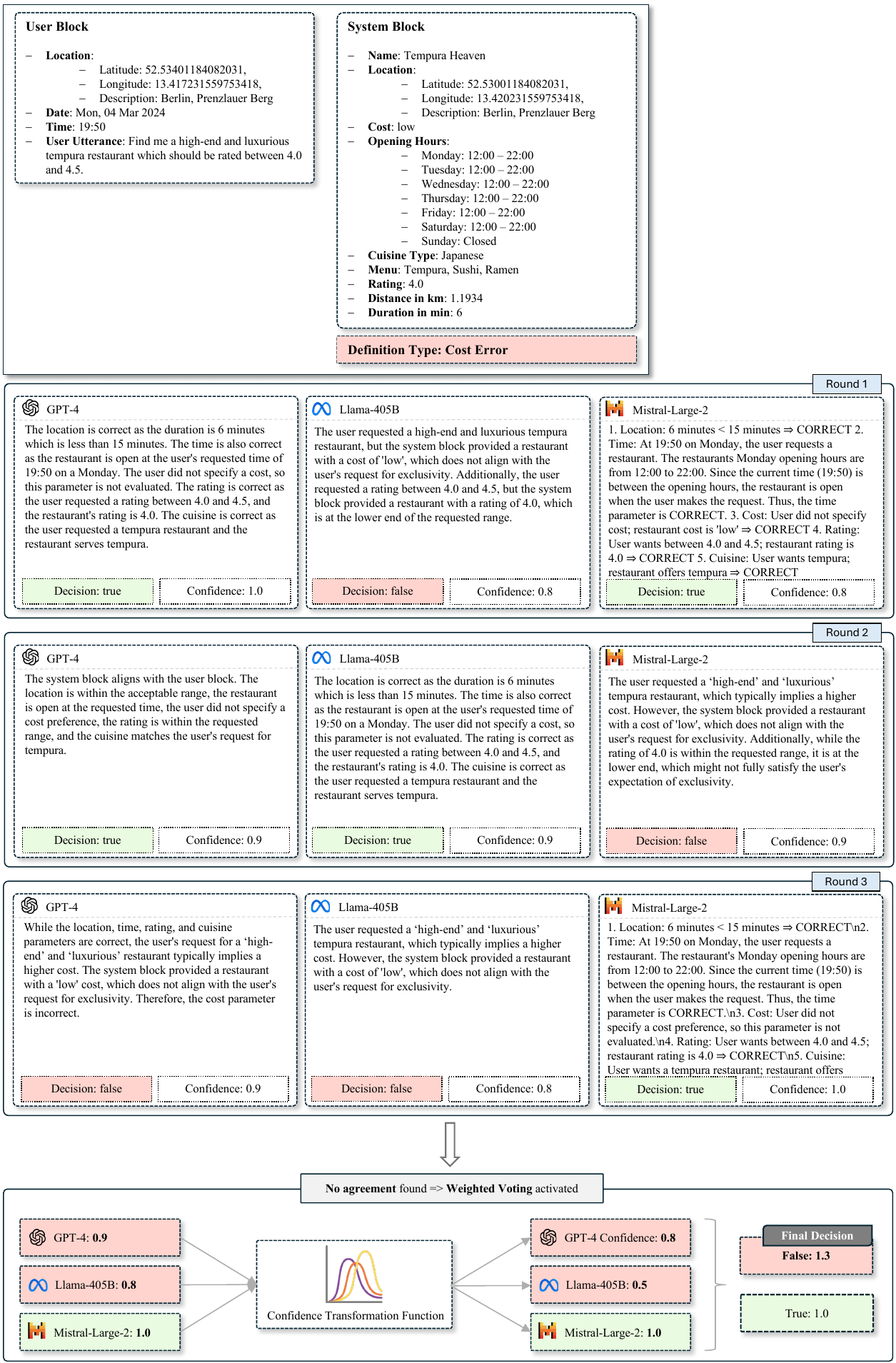}
    \caption{Decision Making Process of Multi-Agent Roundtable}
    \label{fig:AR_discussion_CU}
\end{figure}

\begin{landscape}
\section{Results}
\label{sec:appendix-results}
\begin{longtable}{@{\extracolsep{\fill}}ccccc||ccccc@{}}
    \caption{Precision, Recall, and F1-Score for Context Understanding}\\
    \hline
    \textbf{LLM} & \textbf{Method} & \textbf{Precision} & \textbf{Recall} & \textbf{F1-Score} & \textbf{LLM} & \textbf{Method} & \textbf{Precision} & \textbf{Recall} & \textbf{F1-Score} \\
    \hline
    \endfirsthead
    \hline
    \textbf{LLM} & \textbf{Method} & \textbf{Precision} & \textbf{Recall} & \textbf{F1-Score} & \textbf{LLM} & \textbf{Method} & \textbf{Precision} & \textbf{Recall} & \textbf{F1-Score} \\
    \hline
    \endhead
    \multirow{8}{*}{Llama-8B} & I/O    & 0.275 & \textbf{1.000} & 0.431 & \multirow{8}{*}{GPT-3.5 Turbo} & I/O    & 0.306 & \textbf{0.960} & 0.464 \\*
                & CoT-1  & 0.205 & \textbf{1.000} & 0.340 &                                 & CoT-1  & 0.426 & 0.200 & 0.272 \\*
                & CoT-3  & 0.568 & 0.212 & 0.309 & & CoT-3  & 0.369 & 0.410 & 0.389 \\*
                & CoT-5  & 0.563 & 0.490 & 0.524 & & CoT-5  & 0.418 & \textbf{0.690} & \textbf{0.521} \\*
                & SC-3   & 0.563 & 0.490 & 0.524 & & SC-3   & 0.411 & \textbf{0.690} & 0.515 \\*
                & SC-5   & \textbf{0.575} & 0.500 & 0.535 & & SC-5   & 0.413 & \textbf{0.690} & 0.517 \\*
                & MAB    & 0.386 & 0.930 & 0.545 & & MAB    & 0.489 & 0.230 & 0.313 \\*
                & MAD    & 0.481 & 0.750 & \textbf{0.586} & & MAD    & \textbf{0.500} & 0.120 & 0.194 \\
    \hline
    \multirow{8}{*}{Mistral-Nemo} & I/O    & 0.561 & 0.640 & 0.598 & \multirow{8}{*}{Mistral-Large-2} & I/O    & 0.683 & \textbf{0.990} & 0.808 \\*
                                  & CoT-1  & \textbf{0.731} & 0.190 & 0.302 &                                 & CoT-1  & 0.877 & 0.930 & 0.903 \\*
                                  & CoT-3  & 0.573 & 0.710 & 0.634 &                                 & CoT-3  & 0.869 & 0.930 & 0.899 \\*
                                  & CoT-5  & 0.563 & 0.720 & 0.632 &                                 & CoT-5  & 0.858 & 0.970 & 0.911 \\*
                                  & SC-3   & 0.485 & \textbf{0.830} & 0.613 &                                 & SC-3   & 0.846 & \textbf{0.990} & 0.912 \\*
                                  & SC-5   & 0.485 & \textbf{0.830} & 0.613 &                                 & SC-5   & 0.846 & \textbf{0.990} & 0.912 \\*
                                  & MAB    & 0.694 & 0.735 & \textbf{0.714} &                                 & MAB    & 0.892 & \textbf{0.990} & \textbf{0.938} \\*
                                  & MAD    & 0.605 & 0.780 & 0.681 &                                 & MAD    & \textbf{0.928} & 0.900 & 0.914 \\
    \hline
    \multirow{8}{*}{Llama-405B} & I/O    & 0.683 & 0.970 & 0.802 & \multirow{8}{*}{GPT-4o} & I/O    & 0.925 & 0.980 & 0.951 \\*
                & CoT-1  & 0.752 & 0.970 & 0.847 & & CoT-1  & 0.934 & \textbf{0.990} & 0.961 \\*
                & CoT-3  & 0.892 & 0.990 & 0.938 & & CoT-3  & \textbf{1.000} & 0.909 & 0.952 \\*
                & CoT-5  & 0.868 & 0.990 & 0.925 & & CoT-5  & 0.925 & \textbf{0.990}  & 0.957 \\*
                & SC-3   & 0.868 & 0.990 & 0.925 & & SC-3   & 0.934 & \textbf{0.990}  & 0.961 \\*
                & SC-5   & 0.868 & 0.990 & 0.925 & & SC-5   & 0.934 & \textbf{0.990}  & 0.961 \\*
                & MAB    & 0.913 & \textbf{0.991} & \textbf{0.950} & & MAB & 0.979 & 0.950 & \textbf{0.964} \\*
                & MAD    & \textbf{0.942} & 0.951 & 0.946 & & MAD & 0.979 & 0.920 & 0.948 \\
    \hline
    \multirow{8}{*}{GPT-4 Turbo}   & I/O    & 0.897 & 0.960 & 0.928 & \multirow{8}{*}{GPT-4.1}  & I/O  & 0.933 & \textbf{0.980} & 0.956 \\*
        & CoT-1  & 0.941 & 0.950 & 0.945 & & CoT-1  & 0.939 & 0.930 & 0.935 \\*
        & CoT-3  & 0.912 & 0.930 & 0.921 & & CoT-3 & 0.940 & 0.940 & 0.940 \\*
                  & CoT-5  & 0.883 & \textbf{0.980} & 0.929 & & CoT-5 & 0.938 & 0.900 & 0.918 \\*
                  & SC-3   & 0.941 & 0.950 & 0.945 & & SC-3 & 0.923 & 0.960 & 0.941 \\*
                  & SC-5   & 0.950 & 0.950 & 0.950 & & SC-5 & 0.931 & 0.950 & 0.941 \\*
                  & MAB    & 0.950 & 0.960 & 0.955 & & MAB & \textbf{0.960} & 0.960 & \textbf{0.960} \\*
                  & MAD    & \textbf{0.970} & 0.960 & \textbf{0.965} & & MAD & \textbf{0.960} & 0.950 & 0.955 \\
    \hline
    \multirow{8}{*}{DeepSeek-V3} & I/O    & 0.970 & 0.980 & 0.975 & \multirow{8}{*}{o3-mini} & I/O & 0.970 & 0.980 & 0.975 \\*
                & CoT-1  & 	\textbf{0.980} & 0.960 & 0.970 & & CoT-1  & 0.934 & \textbf{0.990} & 0.961 \\*
                & CoT-3  & \textbf{0.980} & \textbf{0.990} & \textbf{0.985} & & CoT-3  & \textbf{0.980} & \textbf{0.990} & \textbf{0.985} \\*
                & CoT-5  & \textbf{0.980} & 0.960 & 0.970 & & CoT-5  & \textbf{0.980} & 0.960  & 0.970 \\*
                & SC-3   & \textbf{0.980} & \textbf{0.990} & \textbf{0.985} & & SC-3   & 	0.979 & 0.969  & 0.974 \\*
                & SC-5   & 0.979 & 0.969 & 0.974 & & SC-5   & \textbf{0.980} & \textbf{0.990}  & \textbf{0.985} \\*
                & MAB    & \textbf{0.980} & 0.960 & 0.970 & & MAB    & \textbf{0.980} & 0.960 & 0.970 \\*
                & MAD    & 	\textbf{0.980} & 0.970 & 0.975 & & MAD    & \textbf{0.980} & 0.960 & 0.970 \\
    \hline
    \multirow{8}{*}{o3} & I/O    & 0.933 & 0.980 & 0.956 & \multirow{8}{*}{o4-mini} & I/O    & 0.917 & 0.990 & 0.952 \\*
            & CoT-1  & 0.960 & 0.980 & 0.970 & & CoT-1  & \textbf{0.962} & \textbf{1.000} & \textbf{0.980} \\*
            & CoT-3  & 0.961 & \textbf{0.990} & \textbf{0.975} & & CoT-3  & 0.952 & 0.990 & 0.971 \\*
            & CoT-5  & \textbf{0.970} & 0.980 & \textbf{0.975} & & CoT-5  & 0.952 & 0.980  & 0.966 \\*
            & SC-3   & \textbf{0.970} & 0.980 & \textbf{0.975} & & SC-3   & 0.971 & 0.990  & \textbf{0.980} \\*
            & SC-5   & 0.960 & 0.980 & 0.970 & & SC-5   & 0.971 & 0.990  & \textbf{0.980} \\*
            & MAB    & 0.933 & 0.980 & 	0.956 & & MAB    & 0.917 & \textbf{1.000} & 0.957 \\*
            & MAD    & 	0.940 & 0.940 & 0.940 & & MAD    & 0.942 & 0.980 & 0.961 \\
    \hline
    \multirow{8}{*}{DeepSeek-R1}   & I/O    & 0.970 & 0.980 & 0.975 & \multirow{4}{*}{\thead{GPT-3.5 Turbo\\  Llama-8B\\  Mistral-Nemo}}  & \multirow{4}{*}{AR-CoT-5} & \multirow{4}{*}{0.579} & \multirow{4}{*}{0.840} & \multirow{4}{*}{0.686} \\*
        & CoT-1  & 0.990 & 0.990 & \textbf{0.990} & & & & & \\*
        & CoT-3  & 0.980 & 0.990 & 0.985 & & & & & \\*
        & CoT-5  & 0.980 & 0.980 & 0.980 & & & & & \\* \cline{6-10}
        & SC-3   & 0.980 & \textbf{1.000} & \textbf{0.990} & \multirow{4}{*}{\thead{GPT-4 Turbo\\  Mistral-Large-2\\  Llama-405B}}   & \multirow{4}{*}{AR-CoT-5} & \multirow{4}{*}{0.951} & \multirow{4}{*}{0.980} & \multirow{4}{*}{0.966}\\*
        & SC-5   & \textbf{0.990} & 0.990 & \textbf{0.990} & & & & & \\*
        & MAB    & 0.969 & 0.950 & 0.960 & & & & & \\*
        & MAD    & 0.979 & 0.930 & 0.954 & & & & & \\
    \hline
    \multirow{3}{*}{\thead{ o3-mini\\ o4-mini\\ o3}}  & \multirow{3}{*}{AR-CoT-5} & \multirow{3}{*}{0.960} & \multirow{3}{*}{1.000} & \multirow{3}{*}{0.980} & \multirow{3}{*}{\thead{DeepSeek-R1\\  DeepSeek-V3\\  o3-mini}}  & \multirow{3}{*}{AR-CoT-5} & \multirow{3}{*}{0.970} & \multirow{3}{*}{0.990} & \multirow{3}{*}{0.980}\\
    \label{tab:full_results}
\end{longtable}

\clearpage

\begin{figure}
    \centering
    \includegraphics[width=\linewidth]{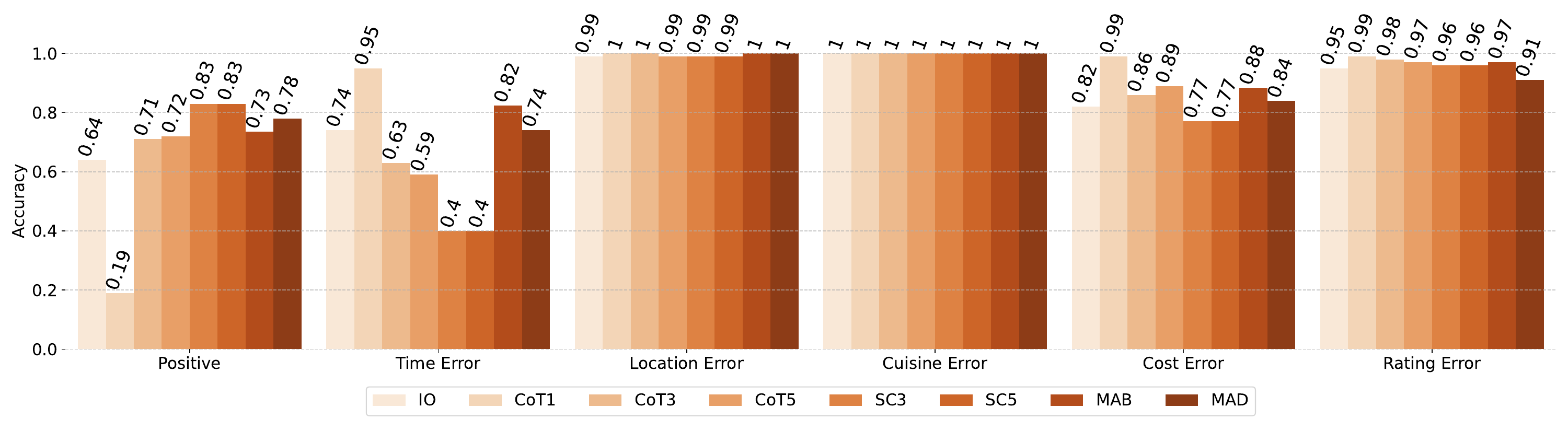}
    \caption{Mistral-Nemo Accuracies for Categories}
    \label{fig:mistral_nemo_accs_cu}
\end{figure}

\begin{figure}
    \centering
    \includegraphics[width=\linewidth]{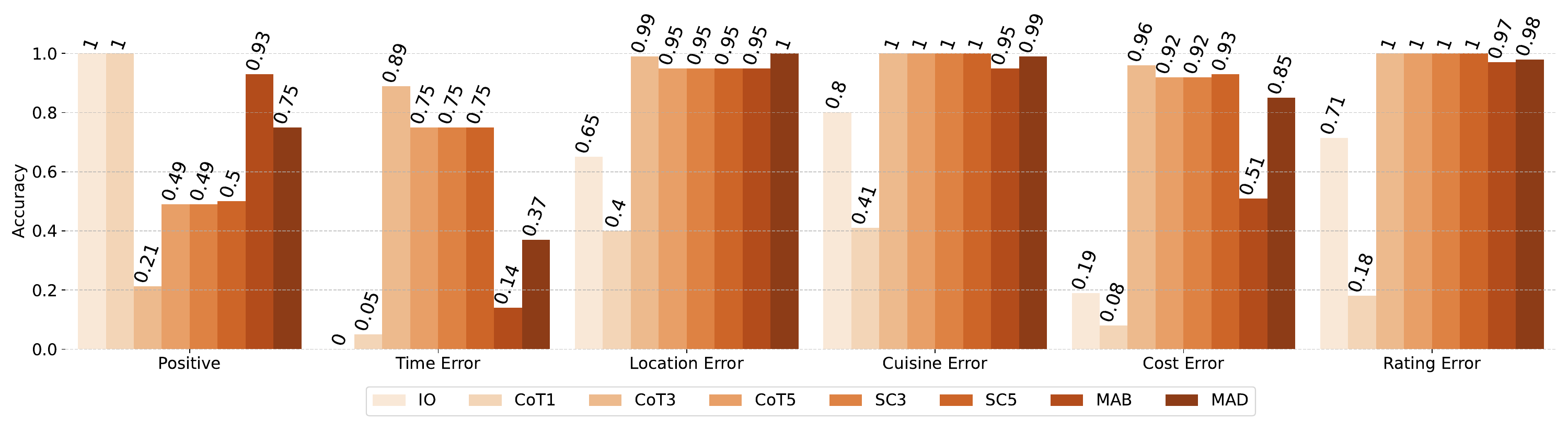}
    \caption{Llama-8B Accuracies for Categories}
    \label{fig:llama_8b_accs_cu}
\end{figure}

\begin{figure}
    \centering
    \includegraphics[width=\linewidth]{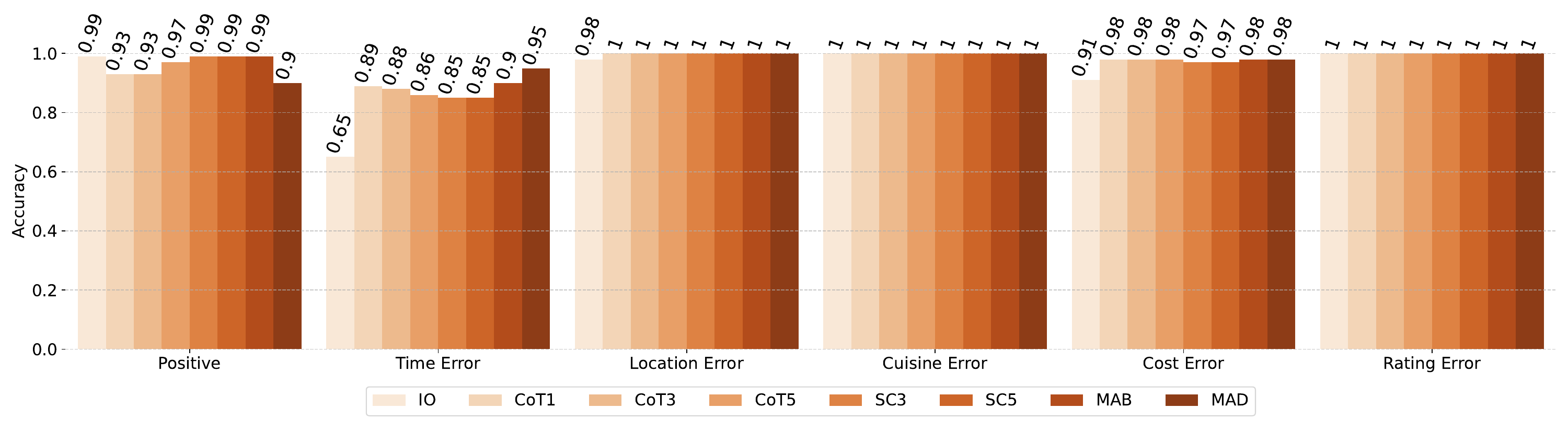}
    \caption{Mistral-Large-2 Accuracies for Categories}
    \label{fig:mistral_large_accs_cu}
\end{figure}

\begin{figure}
    \centering
    \includegraphics[width=\linewidth]{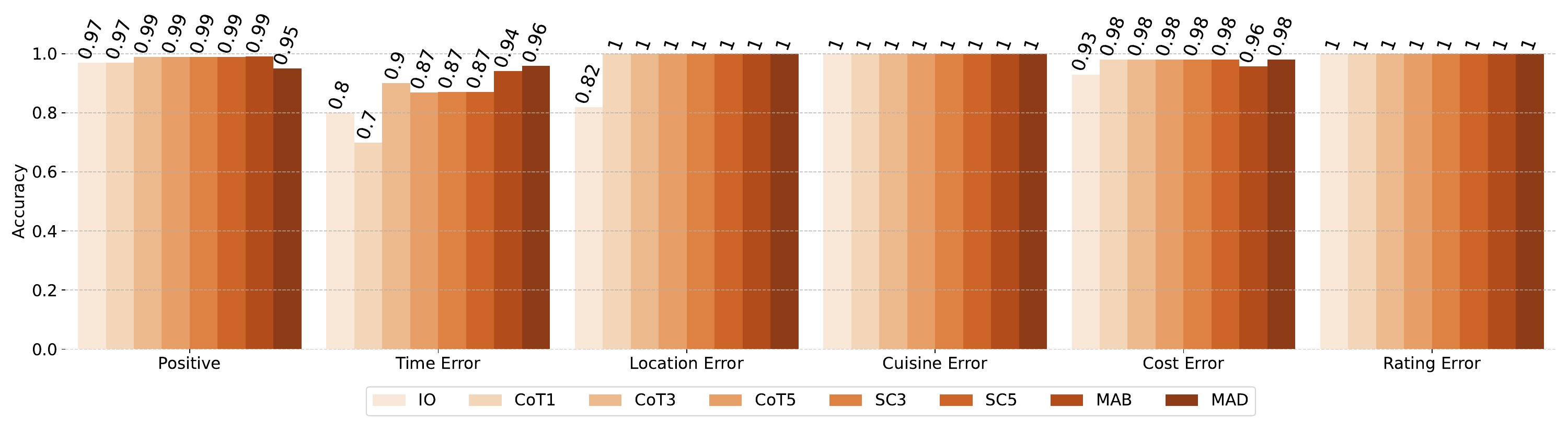}
    \caption{Llama-405B Accuracies for Categories}
    \label{fig:llama-405b_cu_accs}
\end{figure}

\begin{figure}
    \centering
    \includegraphics[width=\linewidth]{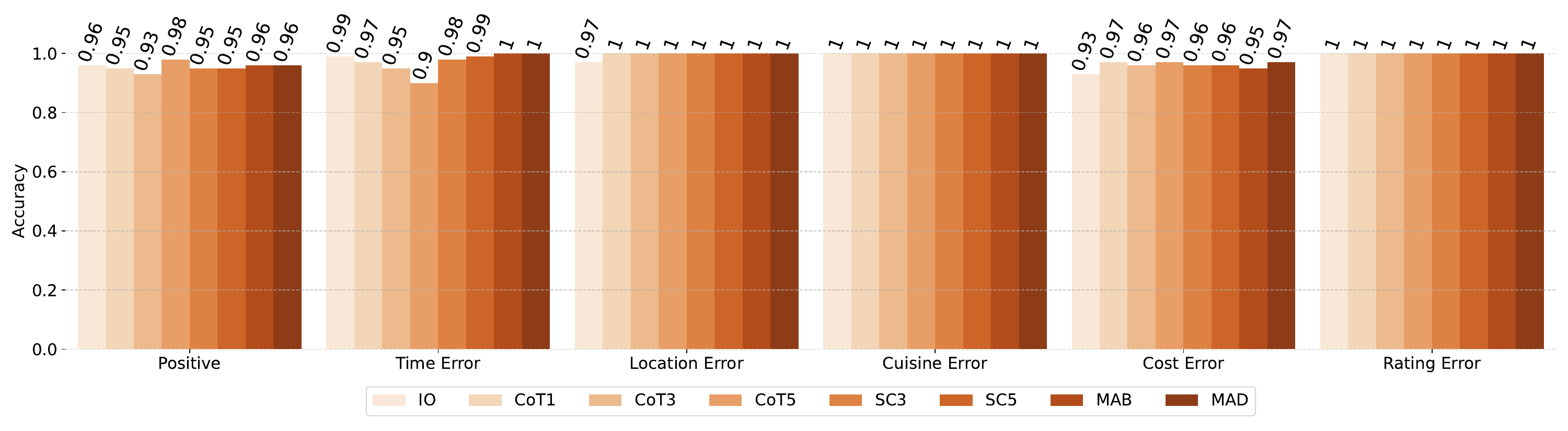}
    \caption{GPT-4 Turbo Accuracies for Categories}
    \label{fig:gpt_4_accs_cu}
\end{figure}

\begin{figure}
    \centering
    \includegraphics[width=\linewidth]{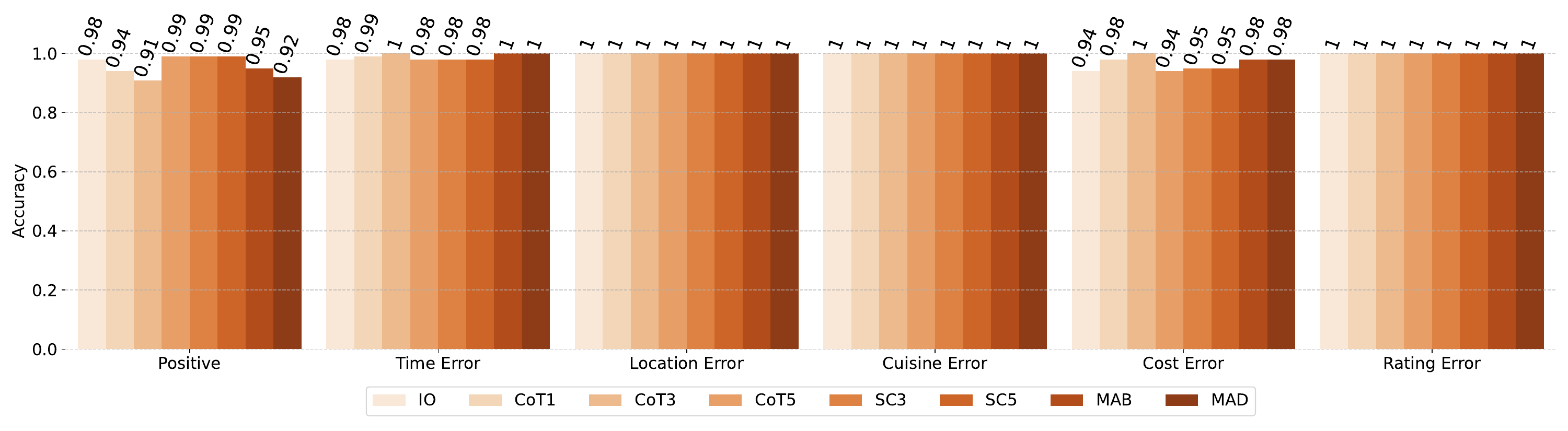}
    \caption{GPT-4o Accuracies for Categories}
    \label{fig:gpt_4o_accs_cu}
\end{figure}

\begin{figure}
    \centering
    \includegraphics[width=\linewidth]{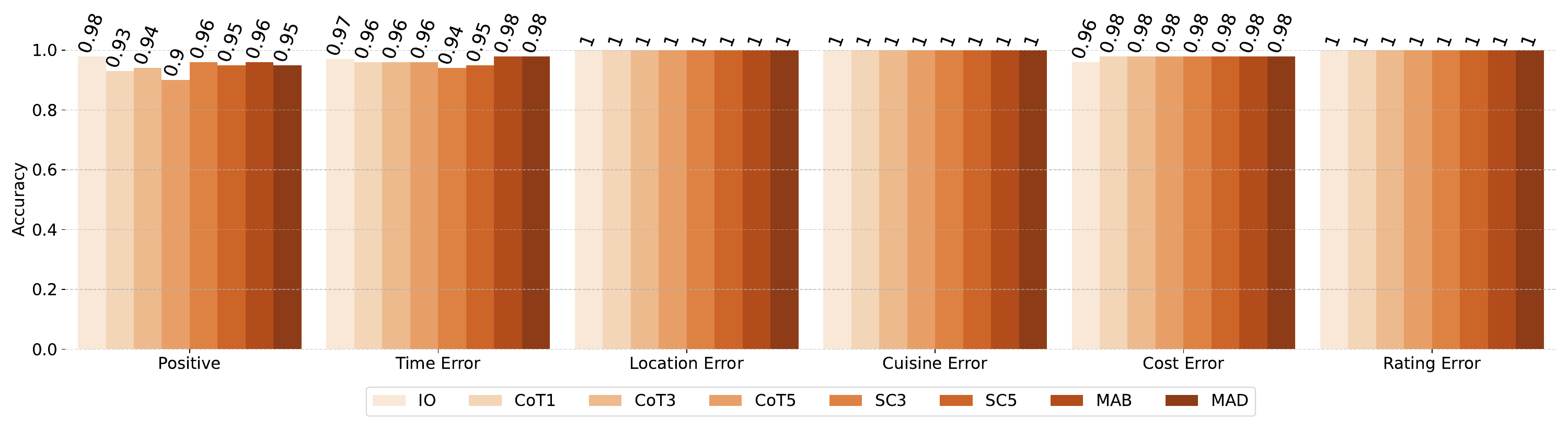}
    \caption{GPT-4 Turbo Accuracies for Categories}
    \label{fig:gpt_4.1_accs_cu}
\end{figure}

\begin{figure}
    \centering
    \includegraphics[width=\linewidth]{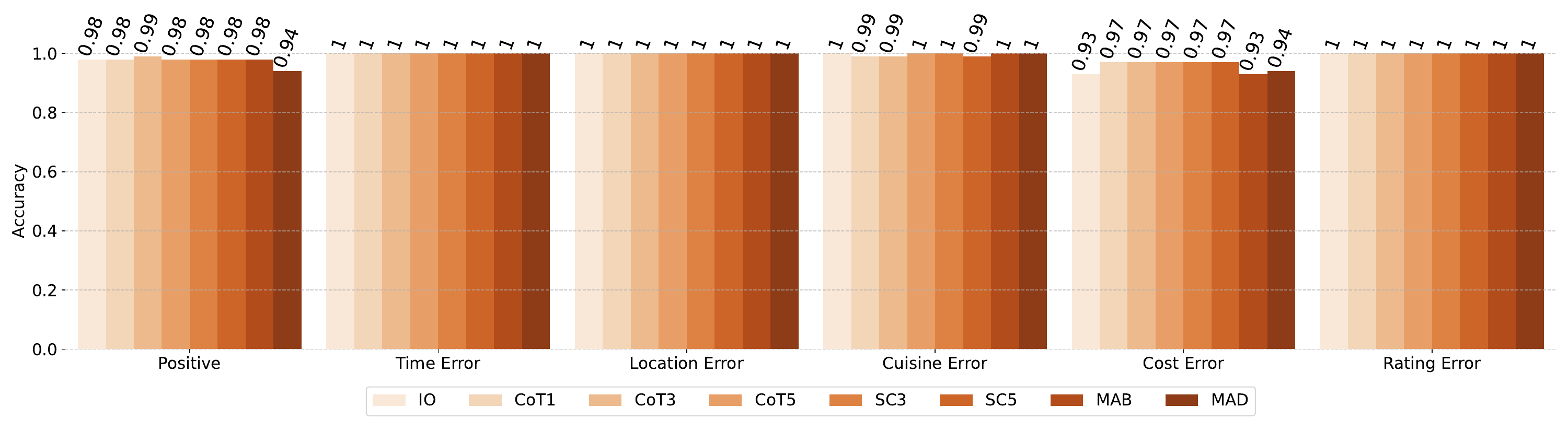}
    \caption{o3 Accuracies for Categories}
    \label{fig:o3_accs_cu}
\end{figure}

\begin{figure}
    \centering
    \includegraphics[width=\linewidth]{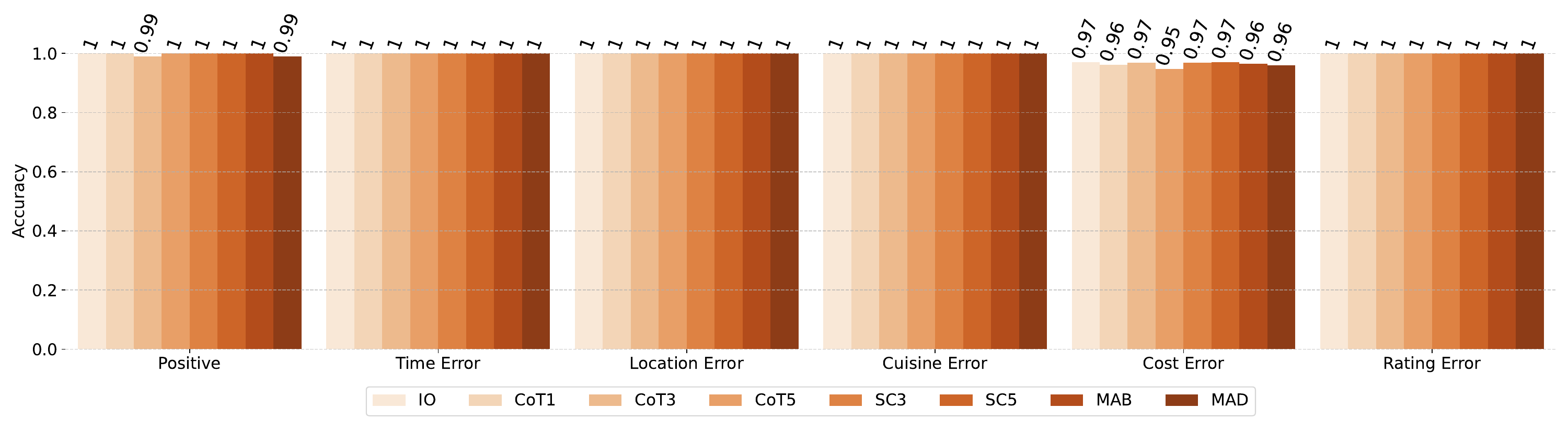}
    \caption{o3-mini Accuracies for Categories}
    \label{fig:o3_mini_accs_cu}
\end{figure}

\begin{figure}
    \centering
    \includegraphics[width=\linewidth]{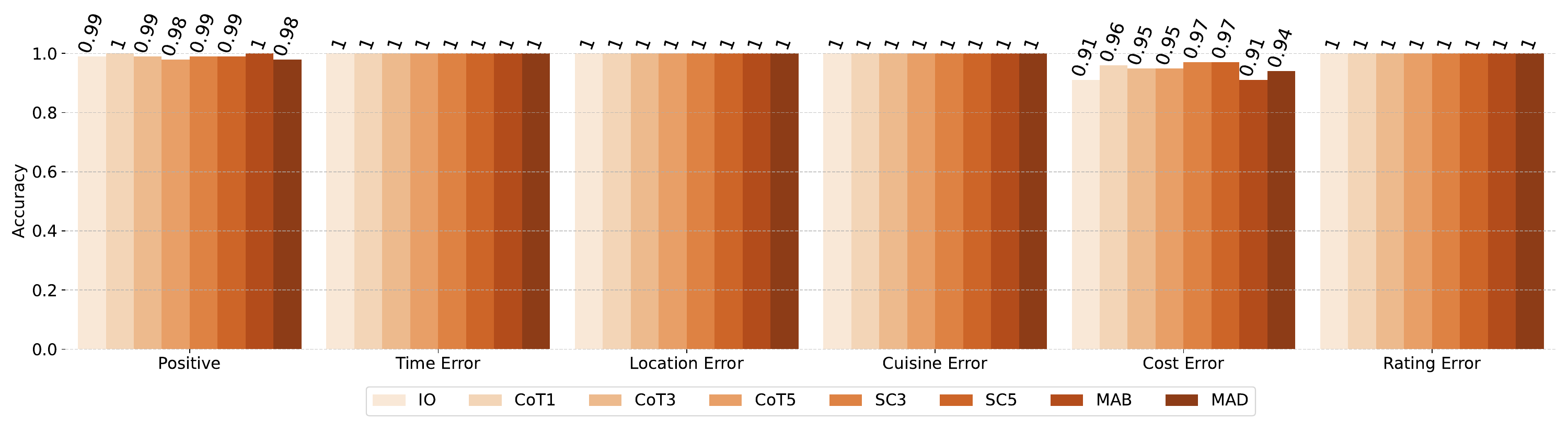}
    \caption{o4-mini Accuracies for Categories}
    \label{fig:o4_mini_accs_cu}
\end{figure}

\end{landscape}

\clearpage

\begin{figure}[!htbp]
    \centering
    \includegraphics[width=\textwidth]{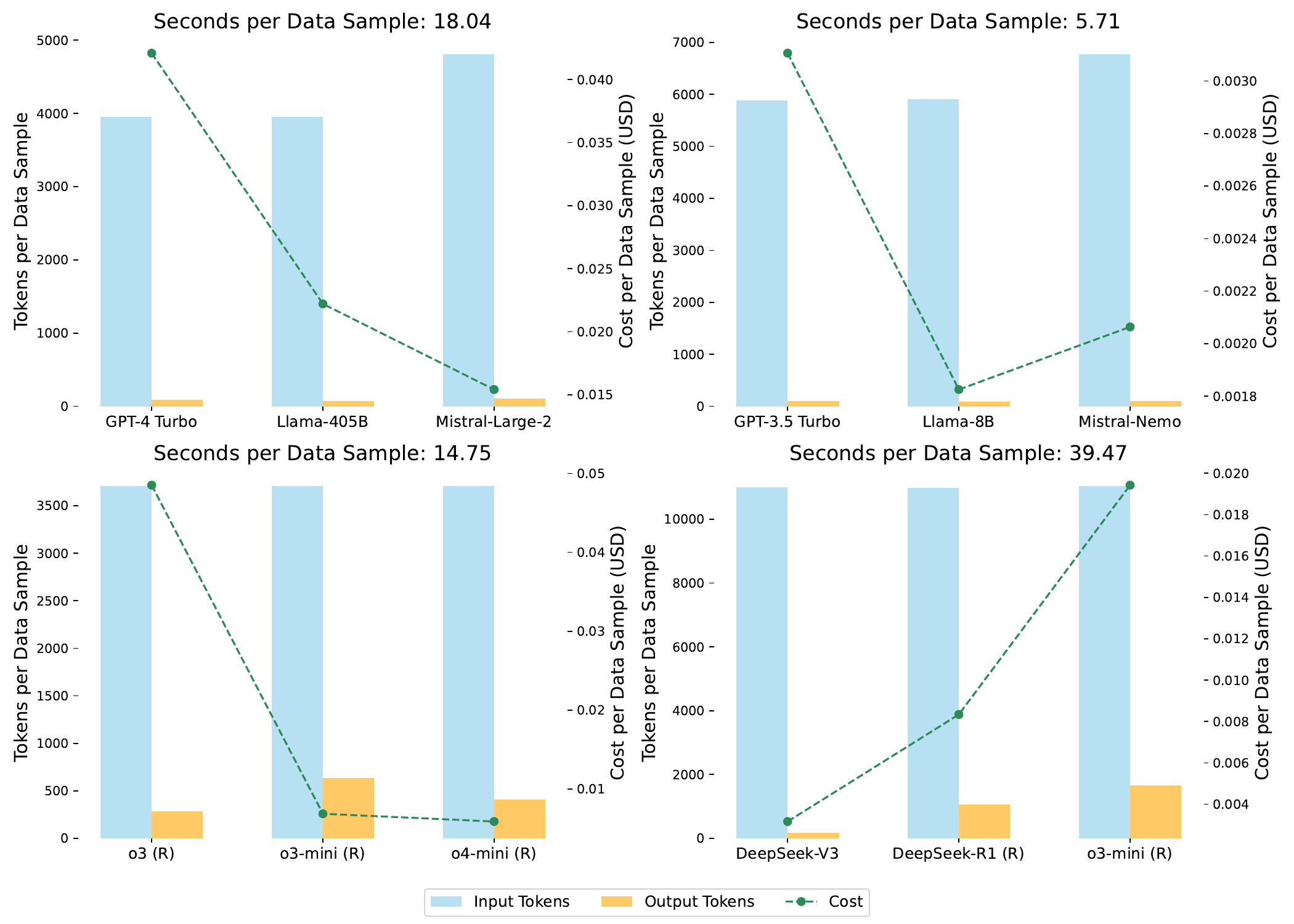}
    \caption{Multi-Agent Roundtable Tokens}
    \label{fig:cu_ar_eff_cost}
\end{figure}

\clearpage
\begin{figure}[!htbp]
    \centering
    \includegraphics[width=1.05\textwidth]{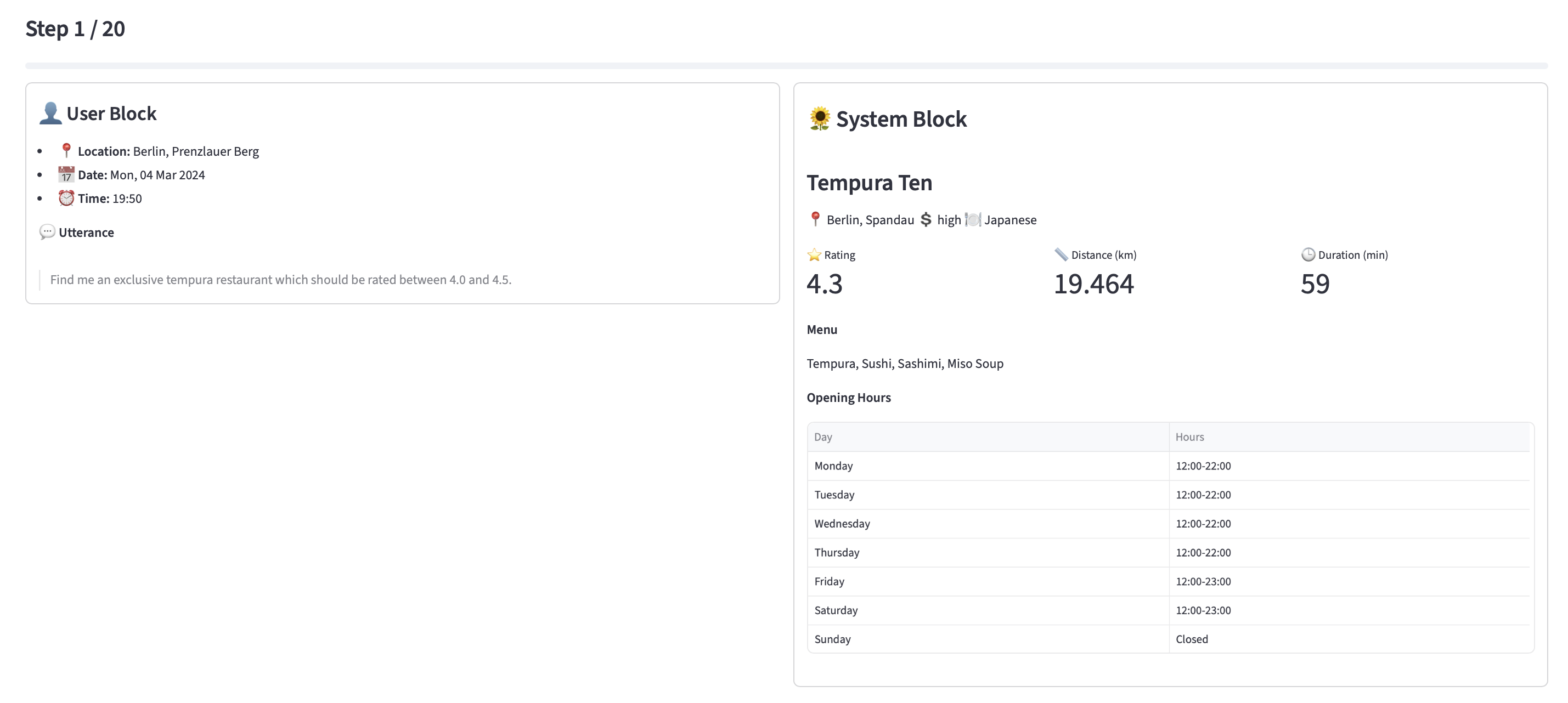}
    \caption{Visualization of the user interface showing user and system blocks for the human agreement study (RQ0)}
    \label{fig:user-system-block}
\end{figure}

\begin{figure}[!htbp]
    \centering
    \includegraphics[width=1.05\textwidth]{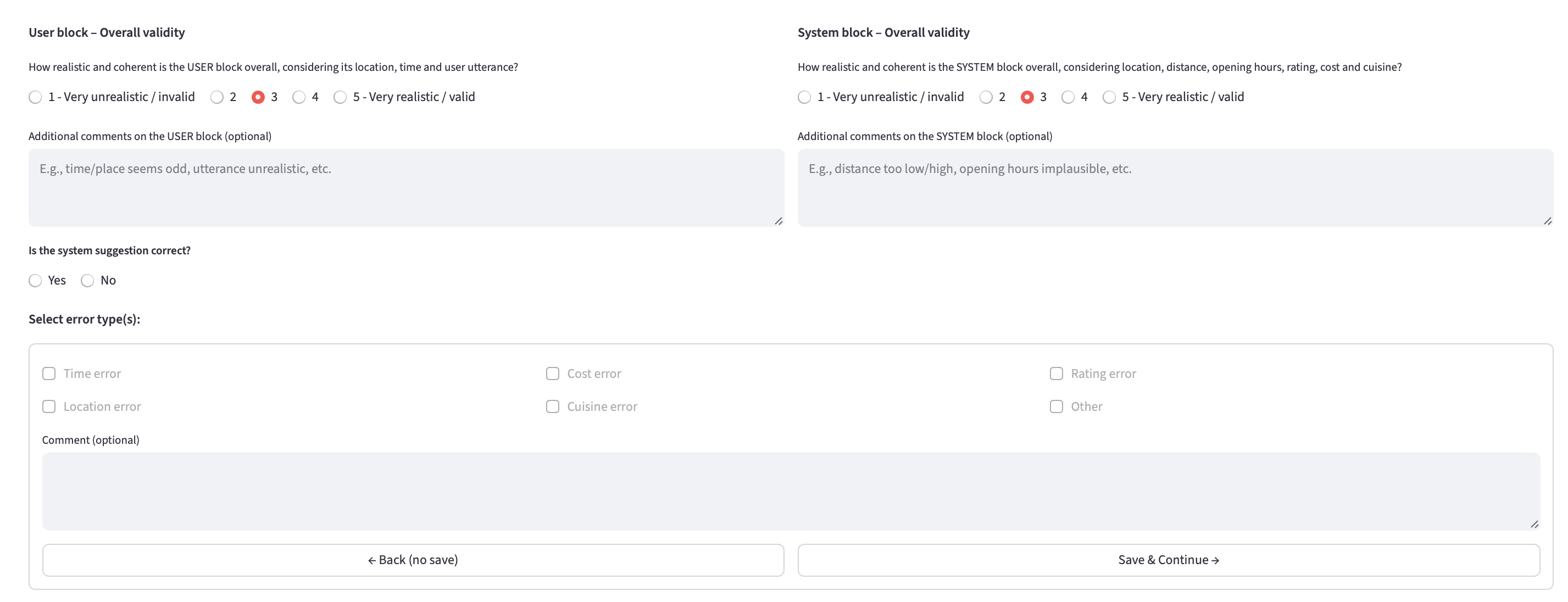}
    \caption{Visualization of the user interface showing questions for the human agreement study (RQ0)}
    \label{fig:user-system-block-questions}
\end{figure}

\end{document}